%% file: main.tex
\documentclass[runningheads]{llncs}

\usepackage{eccv}

\usepackage{eccvabbrv}

\usepackage{graphicx}
\usepackage{booktabs}
\usepackage{pifont}
\newcommand{\cmark}{\textcolor{ForestGreen}{\ding{51}}}
\newcommand{\xmark}{\textcolor{BrickRed}{\ding{55}}}

\usepackage[utf8]{inputenc}
\usepackage[T1]{fontenc}
\usepackage{amsmath, amsfonts}
\usepackage{nicefrac}
\usepackage{microtype}
\usepackage{booktabs, multirow, colortbl}
\usepackage[dvipsnames,table]{xcolor}
\usepackage{capt-of}
\usepackage{float}
\usepackage{cuted}

\usepackage{wrapfig}

\usepackage{xcolor}
\usepackage{pifont}

\usepackage{algorithm}
\usepackage{algpseudocode}

\algrenewcommand\algorithmicrequire{\textbf{Input:}}
\algrenewcommand\algorithmicensure{\textbf{Output:}}
\algrenewcommand\algorithmiccomment[1]{\hfill$\triangleright$~#1}



\newcommand{\jy}[1]{\textcolor{black}{#1}}
\newcommand{\jyn}[1]{\textcolor{black}{#1}}

\definecolor{best}{HTML}{F4C7C3}
\definecolor{second}{HTML}{FFF2CC}
\definecolor{third}{HTML}{E5D5FF}

\newcommand{\Ldep}{\mathcal{L}_{\text{depth}}}
\newcommand{\LnorS}{\mathcal{L}_{\text{normal}}^{\text{SDF}}}
\newcommand{\LnorG}{\mathcal{L}_{\text{normal}}^{\text{3DGS}}}
\newcommand{\Lpbr}{\mathcal{L}_{\text{PBR}}}
\newcommand{\Lphoto}{\mathcal{L}_{\text{image}}}

\definecolor{BoxBlue}{HTML}{0F9ED5}
\definecolor{BoxGreen}{HTML}{4EA72E}
\definecolor{BoxOrange}{HTML}{E97132}

\newcommand{\BlueNum}[1]{\textcolor{BoxBlue}{#1}}
\newcommand{\GreenNum}[1]{\textcolor{BoxGreen}{#1}}
\newcommand{\OrangeNum}[1]{\textcolor{BoxOrange}{#1}}

\usepackage[accsupp]{axessibility}

\usepackage{hyperref}

\renewcommand{\refname}{References}

\begin{document}

\title{\texorpdfstring{COREA: Coupled Relightable 3D Gaussians \\
and SDFs for Efficient Normal Alignment}{COREA: Coupled Relightable 3D Gaussians and SDFs for Efficient Normal Alignment}}

\titlerunning{COREA: Coupled Relightable 3D Gaussians and SDFs}

\author{
Jaeyoon Lee$^{\star}$ \and
Hojoon Jung$^{\star}$ \and
Sungtae Hwang \and
Jihyong Oh$^{\dagger}$ \and
Jongwon Choi$^{\dagger}$
}

\authorrunning{J. Lee et al.}

\institute{
Chung-Ang University, Seoul, Korea \\
\email{\{leejaeyoon, hjjung, sthwang\}@vilab.cau.ac.kr, \{jihyongoh, choijw\}@cau.ac.kr} \\
$^{\star}$Equal contribution. \quad $^{\dagger}$Corresponding authors.
}

\maketitle

\vspace{-1.0em}
\begin{center}
\small \url{https://cau-vilab.github.io/COREA/}
\end{center}
\vspace{-1.0em}

\input{figs/teaser}
\input{sec/0_abstract}
\input{sec/1_intro}
\input{sec/2_related}
\input{sec/3_method}

\input{sec/4_experiments}

\input{sec/5_ablation}

\input{sec/6_conclusion}

{
   \small
   \bibliographystyle{splncs04}
   \bibliography{main}
}
\clearpage
\appendix
\input{sec/X_suppl}

\end{document}

%% file: figs/teaser.tex
\begin{center}
    \includegraphics[width=0.98\linewidth]{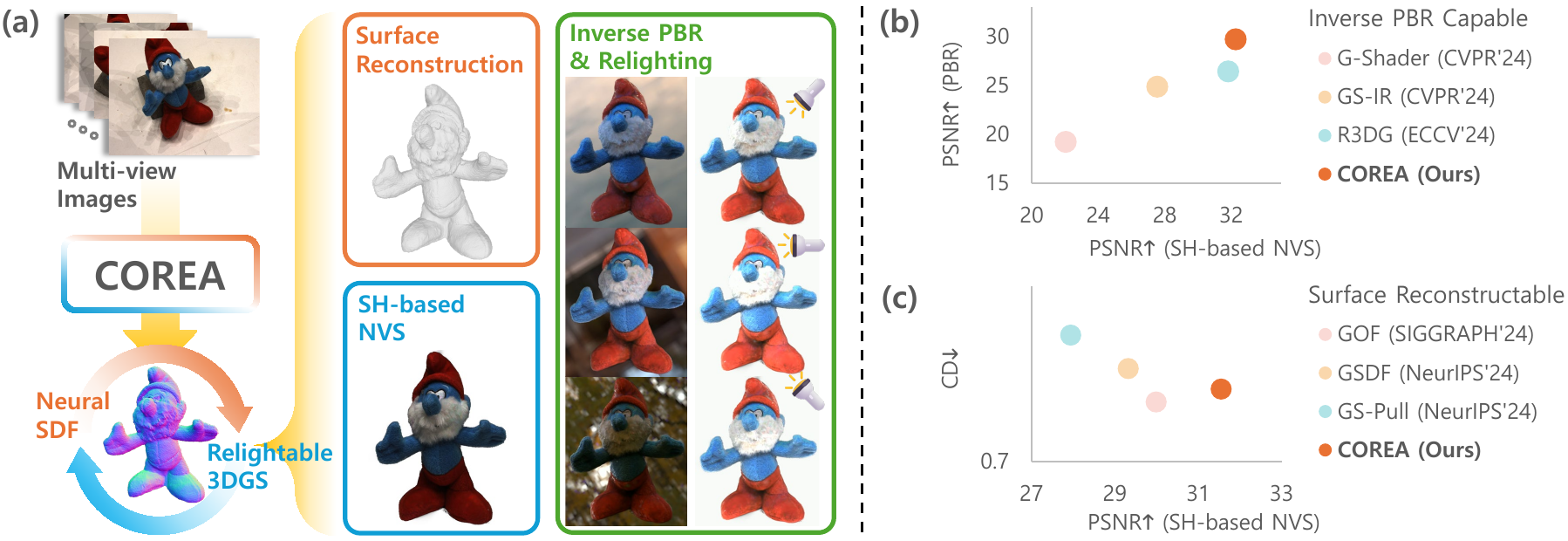}
    \captionof{figure}{
        \textbf{Overview of COREA.}
        \jyn{(a) COREA is the first unified three-tasks framework that couples an SDF and relightable 3D Gaussians on a shared underlying surface for SH-based NVS, surface reconstruction, and inverse PBR.}
        \jyn{(b) Quantitative comparison among inverse-PBR methods shows that COREA achieves strong performance for both SH-based NVS and inverse PBR.}
        \jyn{(c) Among surfaces reconstruction methods, COREA delivers competitive geometric accuracy while maintaining high rendering fidelity.}
    }
    \label{fig:teaser}
    \vspace{-3mm}
\end{center}

%% file: sec/0_abstract.tex
\begin{abstract}
    \jy{
    We present COREA, \jyn{the first unified three-tasks framework that couples an SDF and relightable 3D Gaussians (3DGS) to jointly support SH-based novel-view synthesis (NVS), surface reconstruction, and inverse physically-based rendering (inverse PBR).}
    While recent relightable 3DGS methods have progressed, \jyn{inverse PBR remains bottlenecked by normal estimation, as the discrete nature of 3DGS often yields oversmoothed and unstable normals.}
    \jyn{To address this limitation, COREA couples the complementary geometric properties of an SDF and relightable 3DGS on a shared underlying surface, where geometry-constrained relightable 3DGS provides reliable depth signals to anchor SDF geometry and the continuous SDF normal field provides spatially consistent supervision for Gaussian normal learning.
    We couple these signals through depth-guided alignment and normal supervision with normal-aware densification, and introduce Dual-Density Control to regulate densification by balancing photometric and geometric gradients for stable, memory-efficient training.}
    Experiments on standard benchmarks show that \jyn{COREA is the only framework that supports all three tasks, achieving competitive performance overall, with particularly superior results in \jyn{inverse PBR}.}}
    
    \keywords{3D Gaussian Splatting \and Signed Distance Field \and Inverse Physically-Based Rendering}
\end{abstract}

%% file: sec/1_intro.tex
\section{Introduction}
\label{sec:intro}
\jyn{Recent advancements in 3D Gaussian splatting (3DGS)~\cite{3DGS} enable high-quality, real-time rendering with explicit 3D primitives, which are increasingly integrated with meshes for editing and animation, e.g., via mesh-bound Gaussians in tools such as Blender~\cite{Sugar, Frosting, ManiGS, GaMeS, MaGS, MeshSplats}.
For practical deployment under diverse lighting, 3DGS-based frameworks should jointly support SH-based novel-view synthesis (NVS), surface reconstruction, and inverse physically-based rendering (inverse PBR), so that shading variations induced by manipulation-driven geometry changes are faithfully reproduced.}

\jy{In practice, inverse PBR with 3DGS remains bottlenecked by accurate normal estimation, since surface normals directly govern BRDF-lighting decomposition and relighting quality~\cite{GeoSplatting}.}
\jy{Relightable 3DGS pipelines typically rely on rendered normal maps, often obtained via alpha blending~\cite{R3DG, GSROR2, GeoSplatting, GaussianShader, SVGIR, IRGS, GSIR}, but the discrete splat representation oversmooths the rendered normals and suppresses fine-scale surface variations (Fig.~\ref{fig:normal_comparison}~(A)).}
\jyn{To also support surface reconstruction, many pipelines couple Gaussians with an implicit field, such as neural Signed Distance Fields (SDF), as explicit-implicit hybrid methods~\cite{GSDF, GSROR2, GaussianOpacityFields, neusg, 3dgsr, GS_pull}.}
However, most of these hybrids treat the two branches as separate components, without complementary geometric interaction during joint optimization.
Moreover, many hybrid pipelines still supervise the implicit branch using 3DGS-rendered alpha-blended normals, which blurs the implicit branch's continuous normal field and weakens fine-scale geometric supervision back to the 3DGS branch.
In contrast, pixel-wise depth gradients from the 3DGS depth map preserve sharper local structure, providing a stronger signal for aligning SDF normals (Fig.~\ref{fig:normal_comparison}~(B)).
\jyn{As a result, existing relightable and hybrid pipelines remain incomplete, and none supports all three tasks within a single framework (Tab.~\ref{tab:capability}).}

\jyn{To address these issues, we propose COREA, the first unified three-tasks framework that jointly learns an SDF and relightable 3D Gaussians, uniquely supporting (i) SH-based NVS, (ii) surface reconstruction, and (iii) inverse PBR.}
COREA couples an SDF and relightable 3DGS to use complementary geometric properties for stable normal learning on a shared underlying surface.
The continuous SDF normal field stabilizes Gaussian normal formation, while geometry-constrained relightable 3DGS depth signals refine the SDF geometry.
\jyn{For SDF learning, geometry-constrained relightable 3DGS provides reliable depth signals for SDF ray sampling, and COREA further refines the SDF by aligning SDF normals with pixel-wise depth gradients from the 3DGS depth map rather than alpha-blended normals.}
For Gaussian learning, COREA projects Gaussians onto the SDF surface, aligns Gaussian normals with the SDF normal field, and drives normal-aware densification based on normal inconsistency to capture fine-scale detail.
To keep \jyn{normal-aware densification} stable and efficient, we introduce \jyn{Dual-Density Control (DDC)} that balances \jyn{photometric and geometric gradients}, preventing excessive splitting while preserving geometric fidelity and stable rendering.
With geometrically aligned Gaussians and stabilized normals, we perform inverse PBR~\cite{Rendering_equation} to robustly disentangle BRDF parameters and lighting, enabling faithful relighting under novel lighting.

As shown in Fig.~\ref{fig:teaser}~(a), COREA is \jyn{the only unified three-tasks framework that simultaneously supports SH-based NVS, surface reconstruction, and inverse PBR (Tab.~\ref{tab:capability}) by coupling relightable 3DGS and SDF to use complementary geometric properties on a shared underlying surface.}
It achieves superior performance in \jyn{SH-based NVS and inverse PBR} (Fig.~\ref{fig:teaser}~(b)), while also delivering high-quality \jyn{reconstructed surfaces} (Fig.~\ref{fig:teaser}~(c)).

Our key contributions are as follows:
\begin{itemize}
    \setlength{\itemsep}{0pt}%
    \setlength{\parskip}{0pt}%
    \item \jyn{We present COREA, the first unified three-tasks framework that jointly learns an SDF and relightable 3DGS, and supports SH-based NVS, surface reconstruction, and inverse PBR.}
    \item \jyn{We couple an SDF and relightable 3DGS on a shared underlying surface: geometry-constrained relightable 3DGS provides reliable supervision for SDF geometry learning, and the continuous SDF normal field stabilizes Gaussian normal formation.}
    \item \jyn{We propose a dual-density control mechanism that regularizes densification using both photometric and geometric gradients, preventing excessive Gaussian splitting while maintaining geometric fidelity and stable rendering.}
    \item \jyn{Experiments show that COREA achieves superior performance in SH-based NVS and inverse PBR, while delivering high-quality reconstructed surfaces on diverse benchmarks.}
\end{itemize}

\input{figs/LR/checkbox_normal}

%% file: figs/LR/checkbox_normal.tex
\setlength{\textfloatsep}{10pt}
\begin{figure}[!t]
\centering

\begin{minipage}[t]{0.49\linewidth}
\centering
\captionsetup{type=table,font=small,skip=3pt,belowskip=0pt}
\caption{
\textbf{Task Support Comparison.}
Comparison of supported tasks: SH-based NVS (SH), Surface Reconstruction (SR), and Inverse PBR (PBR).
}
\renewcommand{\arraystretch}{1.05}
\setlength{\tabcolsep}{4pt}
\resizebox{\linewidth}{!}{%
\begin{tabular}{l|l|ccc}
\toprule
\textbf{Category} & \textbf{Method} & \textbf{SH} & \textbf{SR} & \textbf{PBR} \\
\midrule

Vanilla
& 3DGS (SIGGRAPH’23)         & \textcolor{green}\cmark & \textcolor{red}\xmark & \textcolor{red}\xmark \\

\midrule

\multirow{4}{*}{Relightable}
& GaussianShader (CVPR’24)   & \textcolor{green}\cmark & \textcolor{red}\xmark & \textcolor{green}\cmark \\
& GS-IR (CVPR’24)            & \textcolor{green}\cmark & \textcolor{red}\xmark & \textcolor{green}\cmark \\
& R3DG (ECCV’24)             & \textcolor{green}\cmark & \textcolor{red}\xmark & \textcolor{green}\cmark \\
& SVG-IR (CVPR’25)           & \textcolor{green}\cmark & \textcolor{red}\xmark & \textcolor{green}\cmark \\

\midrule

\multirow{5}{*}{Hybrid}
& GOF (SIGGRAPH’24)          & \textcolor{green}\cmark & \textcolor{green}\cmark & \textcolor{red}\xmark \\
& GSDF (NeurIPS’24)          & \textcolor{green}\cmark & \textcolor{green}\cmark & \textcolor{red}\xmark \\
& GS-Pull (NeurIPS’24)       & \textcolor{green}\cmark & \textcolor{green}\cmark & \textcolor{red}\xmark \\
& GS-ROR$^{2}$ (TOG’25)      & \textcolor{red}\xmark & \textcolor{green}\cmark & \textcolor{green}\cmark \\
& \textbf{COREA (Ours)}      & \textcolor{green}\cmark & \textcolor{green}\cmark & \textcolor{green}\cmark \\

\bottomrule
\end{tabular}
}
\label{tab:capability}
\end{minipage}
\hfill
\begin{minipage}[t]{0.49\linewidth}
\vspace{0pt}
\centering
\includegraphics[width=\linewidth]{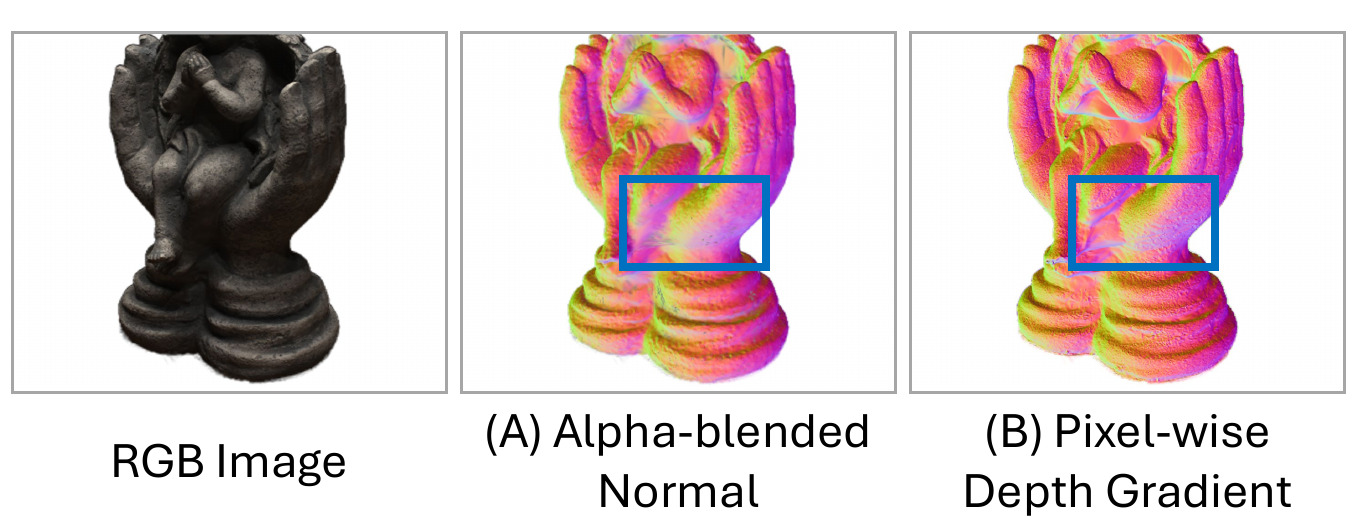}
\captionsetup{type=figure,font=small,skip=3pt,belowskip=0pt}
\caption{
\textbf{Comparison of normal supervision signals at an intermediate training stage (35k iterations).} 
(A) Alpha-blended normals are blurrier and noisier, providing unstable supervision for SDF normal learning.
(B) Pixel-wise depth gradients preserve sharper structure, yielding more stable SDF alignment.
}
\label{fig:normal_comparison}
\end{minipage}

\end{figure}

%% file: sec/2_related.tex
\section{Related Work}

\subsection{Inverse Physically-based Rendering}
Inverse PBR aims to recover accurate geometry, material properties, and illumination by modeling light-surface interactions from multi-view observations and disentangling reflectance from lighting.
NeRF~\cite{NeRF}-based inverse rendering approaches~\cite{NeILF, NeILFpp, LightSampling, NeRD, NeFII, NeRFactor} jointly estimate geometry, reflectance, and lighting, but typically require expensive volumetric rendering and remain computationally demanding.
Several reflectance-aware works~\cite{RefNeRF, IndirectNeRF} further improve appearance decomposition by modeling specular reflections and indirect illumination.

Recent studies extend 3DGS to inverse rendering by building on its efficient explicit representation.
Methods such as GIR~\cite{GIR}, GS-IR~\cite{GSIR}, and GaussianShader~\cite{GaussianShader} incorporate inverse rendering formulations into Gaussian representations by factorizing geometry, material, and lighting, enforcing normal regularization, and learning shading functions for reflective scenes.
R3DG~\cite{R3DG} introduces point-based ray tracing into 3DGS to enable finer BRDF and lighting decomposition.
IRGS~\cite{IRGS} incorporates the full rendering equation into 2D Gaussian Splatting through differentiable Gaussian ray tracing.
Ref-Gaussian~\cite{RefGaussian} captures fine inter-reflection effects using a physically based deferred rendering pipeline, while RNG~\cite{RNG} replaces the analytical rendering equation with neural modules that learn Gaussian-light interactions.
GeoSplatting~\cite{GeoSplatting} and SVG-IR~\cite{SVGIR} further incorporate geometry-aware optimization and structured parameter disentanglement to model geometry, material, and illumination within Gaussian representations.

However, Gaussian normals rendered via alpha blending are often oversmoothed due to the discrete splat representation, making BRDF-lighting decomposition unreliable for inverse PBR.
To address this limitation, we propose a unified framework that couples an SDF and relightable 3DGS, enabling complementary geometric supervision during optimization.
This coupling provides stable surface normals and supports faithful BRDF and lighting decomposition under complex illumination.

\subsection{Hybrid 3DGS Methods for Surface Reconstruction}
Recent explicit-implicit hybrid studies integrate 3DGS with neural implicit representations such as SDF to improve geometric stability and surface reconstruction~\cite{GSDF, neusg, 3dgsr, GS_pull, GaussianRoom}.
Most hybrid approaches primarily combine the two representations by assigning separate roles: Gaussians provide strong photometric rendering with efficient explicit primitives, while implicit fields supply a continuous geometric representation for surface reconstruction.
GSDF~\cite{GSDF} jointly optimizes Gaussian primitives and an SDF representation to stabilize geometry.
Several works enforce geometric alignment between Gaussian primitives and implicit fields.
Methods such as G2SDF~\cite{G2SDF}, DiGS~\cite{DiGS}, GS-SDF~\cite{GSSDF}, and MonoGSDF~\cite{monoGSDF} align Gaussian centers with the zero-level set of implicit fields or estimate SDF fields from Gaussian distributions.
GS-Pull~\cite{GS_pull} further refines Gaussian geometry by pulling Gaussian centers toward the SDF zero-level set using SDF gradient guidance.
Other approaches improve geometry directly within Gaussian representations.
Gaussian Opacity Fields (GOF)~\cite{GaussianOpacityFields} introduce opacity-driven geometric modeling that stabilizes geometry extraction.
GS-ROR$^2$~\cite{GSROR2} incorporates SDF guidance into a relightable 3DGS framework by pruning Gaussians that deviate from the SDF zero-level set.
However, this role-separated coupling is not designed to induce complementary geometric interaction during joint training.
Consequently, the interaction remains weak and indirect, often relying on auxiliary constraints or pruning operations, which limits geometric co-optimization between the two branches.
To move beyond simple combination, we couple an SDF with geometry-constrained relightable 3DGS and jointly optimize them through mutual geometric supervision.
By combining complementary geometric properties of the SDF normal field and geometry-constrained relightable 3DGS, our strategy enables consistent alignment and stable normals, improving fine-scale detail.

%% file: sec/3_method.tex
\section{Proposed Method: COREA}
\input{figs/frame_work}
\jy{As illustrated in Fig.~\ref{fig:framework}, COREA couples an \jyn{SDF and relightable 3DGS on a shared underlying surface through the continuous SDF normal field and geometry-constrained depth signals from 3D Gaussians.}}
\jyn{The continuous SDF normal field stabilizes Gaussian normal formation and supports normal-aware densification to capture fine-scale geometry.
Conversely, geometry-constrained relightable 3DGS provides reliable depth signals for SDF alignment, while pixel-wise depth gradients from the 3DGS depth map refine SDF geometry by guiding SDF normal alignment.
A Dual-Density Control mechanism regularizes densification by balancing photometric and geometric gradients, stabilizing geometry learning and suppressing redundant splitting.
COREA couples stable geometry learning from coarse depth to fine surface detail on a shared underlying surface and jointly supports SH-based NVS, surface reconstruction, and inverse PBR.}

\input{figs/Algorithm}

\jyn{COREA operates in two stages (Alg.~\ref{alg:corea_train}.)
The first stage focuses on \jyn{geometric coupling between an SDF and relightable 3DGS (Sec.~\ref{BDS}).}
This stage alternates between two complementary steps: 
(i) Depth-guided SDF Alignment (DSA) performs \jyn{depth-guided ray sampling using the relightable 3DGS depth map and refines the SDF normal field by matching SDF normals to pixel-wise depth gradients from the 3DGS depth map.}
(ii) Normal-guided Gaussian Alignment (NGA) projects Gaussians onto the \jyn{SDF surface} and then refines Gaussian normals with the SDF normal field to capture fine-scale geometry and drive normal-aware densification.
Dual Density Control regularizes \jyn{normal-aware densification} with both photometric and geometric gradients to prevent excessive splitting.
After the geometry converges, the second stage performs inverse PBR (Sec.~\ref{PBR}), optimizing BRDF parameters and lighting on the learned geometry to enable accurate and spatially consistent relighting.}
In our implementation, we use R3DG~\cite{R3DG} for the 3D Gaussian representation and NeuS~\cite{neus} for the \mbox{SDF representation}.
\jyn{Further algorithmic details of DSA and NGA are provided in the \textit{supplementary material.}}

\subsection{Geometric Coupling Between SDF and Relightable 3DGS}
\label{BDS}
\jyn{To jointly refine geometry across an SDF and relightable 3DGS, we enforce cross-representation geometric consistency by coupling the two representations through the continuous SDF normal field and 3DGS depth signals.}
This strategy consists of two complementary steps: DSA and NGA.

\subsubsection{Depth-guided SDF Alignment (DSA).}
\label{DSA}
To achieve accurate and stable geometry learning, we design DSA to align \jyn{an SDF with relightable 3DGS.}
\jy{Relightable 3DGS renders depth under a depth-distribution uncertainty constraint~\cite{R3DG}, producing reliable \jyn{depth signals to anchor the coarse SDF geometry}, while pixel-wise depth gradients capture fine local surface variations.}
Since the SDF gradient inherently represents surface normals, we refine fine-scale geometry by aligning \jyn{SDF normals} with depth-gradient directions derived from the 3DGS depth map.

We begin with coarse alignment by rendering a depth map $\mathbf{d}_{\text{3DGS}}$ from relightable 3DGS to guide ray sampling of an SDF.
Following the ray-based sampling strategy of GSDF~\cite{GSDF}, we adopt an adaptive sampling interval centered on $\mathbf{d}_{\text{3DGS}}$, allowing the SDF to focus on learning geometry near the \jyn{3DGS-rendered depth.}
To encourage geometric consistency, we minimize an $\mathcal{L}_1$ loss ($\mathcal{L}_{\text{depth}}$) between the SDF-rendered and 3DGS-rendered depth maps.

\jyn{For fine-grained alignment, we compute the 2D pixel-wise gradients of $\mathbf{d}_{\text{3DGS}}$ (Fig.~\ref{fig:normal_comparison}~(B)), instead of alpha-blended normals that become oversmoothed when compositing discrete Gaussian primitives (Fig.~\ref{fig:normal_comparison}~(A)).}
We define $\mathbf{n}_{\text{3DGS}}^{\text{DG}} =\nabla \mathbf{d}_{\text{3DGS}} \allowbreak / \allowbreak\| \nabla \mathbf{d}_{\text{3DGS}} \|$.
This choice provides a stronger signal for aligning SDF normals, as validated in our ablation study (Tab.~\ref{tab:ablation_normal_comparison}).
Since the SDF implicitly encodes surface normals via its gradient, we minimize a cosine-similarity loss between the normalized SDF gradient $\mathbf{n}_{\text{SDF}}$ and $\mathbf{n}_{\text{3DGS}}^{\text{DG}}$:
{
\setlength{\abovedisplayskip}{10pt}
\setlength{\belowdisplayskip}{10pt}
\begin{equation}
    \mathcal{L}_{\text{normal}}^{\text{SDF}} = \frac{1}{N} \sum_{i=1}^{N} 
    \left( 1 - \left\langle \mathbf{n}_{\text{SDF}}^{(i)}, \mathbf{n}_{\text{3DGS}}^{\text{DG}(i)} \right\rangle \right),
\label{eq:sdf_normal}
\end{equation}
}
where \(N\) is the number of valid pixels.  
The loss $\mathcal{L}_{\text{normal}}^{\text{SDF}}$ encourages the SDF gradient field to follow the depth-gradient directions of the 3DGS depth map under a local planarity assumption.

\jy{This depth-guided SDF learning yields a continuous normal field that later serves as a stable geometric prior for Gaussian normal learning in NGA, strengthening cross-representation geometric coupling beyond using alpha-blended normal signals alone.}
Consequently, the SDF reconstructs a surface tightly aligned with the \jyn{3DGS depth surface}, providing a reliable foundation for subsequent relighting and geometry extraction.

\input{figs/DDC}

\subsubsection{Normal-guided Gaussian Alignment (NGA).}
\label{NGA}
Accurate BRDF decomposition and relighting rely on reliable fine-scale geometry and surface normals\jyn{~\cite{GeoSplatting}.}
\jyn{Since alpha-blended normals from discrete Gaussians are often oversmoothed and unstable in fine regions (Fig.~\ref{fig:normal_comparison}~(A)), we introduce NGA to refine relightable 3DGS on the shared underlying surface by leveraging the continuous SDF normal field as a stable reference for normal learning.}

\jy{We first \jyn{align relightable 3DGS to the SDF} by minimizing the depth consistency loss $\mathcal{L}_{\text{depth}}$ between \jyn{their rendered depth maps.}}
Fine-scale accuracy is then improved by minimizing a pixel-wise cosine-similarity loss between \jyn{the 3DGS-rendered alpha-blended normal map $\mathbf{n}^{\text{AB}}_{\text{3DGS}}$ and the SDF normal map $\mathbf{n}_{\text{SDF}}$}, where $\mathbf{n}^{\text{AB}}_{\text{3DGS}}$ serves as a differentiable coupling interface to update Gaussians rather than a stable supervision signal:
{
\setlength{\abovedisplayskip}{10pt}
\setlength{\belowdisplayskip}{10pt}
\begin{equation}
    \mathcal{L}_{\text{normal}}^{\text{3DGS}} = \frac{1}{N} \sum_{i=1}^{N} \left( 1 - \left\langle \mathbf{n}_{\text{3DGS}}^{\text{AB}(i)}, \mathbf{n}_{\text{SDF}}^{(i)}
    \right\rangle \right).
\label{eq:3DGS_normal}
\end{equation}
}
\jy{This cross-representation alignment distills \jyn{SDF continuity into the discrete Gaussians,} enabling more stable and spatially consistent normal learning.}

To further enhance geometric detail, we use the gradient magnitude of $\mathcal{L}_{\text{normal}}^{\text{3DGS}}$ as an additional criterion for adaptive Gaussian densification.
While the original 3DGS performs densification based only on photometric gradients to improve rendering quality, we additionally densify Gaussians that exhibit large gradients of $\mathcal{L}_{\text{normal}}^{\text{3DGS}}$.
This criterion encourages refinement in regions with large geometric discrepancies, improving both rendering fidelity and fine geometric detail.

\subsection{Dual-Density Control (DDC)}
\label{DDC}
\setlength{\textfloatsep}{6pt}
The original adaptive density control of 3DGS splits a Gaussian when the accumulated gradient magnitude exceeds a predefined threshold.  
In NGA, we densify Gaussians according to the pixel-wise normal alignment loss $\mathcal{L}_{\text{normal}}^{\text{3DGS}}$, which enhances fine-scale geometry.
\jyn{Here, $\mathcal{L}_{\text{normal}}^{\text{3DGS}}$ is driven by the continuous SDF normal field, producing an additional geometric gradient that complements the photometric gradients used in standard 3DGS densification.}
However, since each Gaussian projects onto multiple pixels, the gradients accumulate excessively, often resulting in unnecessary and redundant splits.

To mitigate such redundant splitting, we propose a Dual-Density Control (DDC) strategy, illustrated in Fig.~\ref{fig:DDC}.
Inspired by SteepGS~\cite{wang2025steepest}, DDC jointly considers \jyn{photometric gradients} and \jyn{geometric gradients}, induced by $\mathcal{L}_{\text{image}}^{\text{3DGS}}$ and the SDF-driven normal loss $\mathcal{L}_{\text{normal}}^{\text{3DGS}}$, respectively.
For each Gaussian \( i \), we compute splitting matrices \( S_{\text{image}}^{(i)} \) and \( S_{\text{normal}}^{(i)} \) from the two loss terms, respectively.
These matrices approximate the local curvature of the loss landscape with respect to the Gaussian center \( \mu^{(i)} \), providing both a criterion for whether splitting will reduce the loss and a direction in which to place the split Gaussians.
To balance the influence of both loss terms, we then form a weighted sum of the two matrices:
{
\setlength{\abovedisplayskip}{10pt}
\setlength{\belowdisplayskip}{10pt}
\begin{equation}
    S_{\text{total}}^{(i)} = (1 - \alpha) \cdot S_{\text{image}}^{(i)} + \alpha \cdot S_{\text{normal}}^{(i)},
\label{eq:splitting_total}
\end{equation}
}
where \(\alpha \in [0,1]\) controls the contribution of the normal loss to the splitting decision.
A Gaussian is split only if the minimum eigenvalue of \( S_{\text{total}}^{(i)} \) is negative, i.e., \( \lambda_{\min}(S_{\text{total}}^{(i)}) < 0 \), indicating a descending curvature direction that benefits overall loss reduction.

Through DDC, \jyn{we regulate densification by balancing photometric and geometric gradients, stabilizing convergence between the two representations.}
This curvature-guided splitting effectively constrains excessive Gaussian growth, maintaining geometric fidelity and stable rendering performance.
\jyn{Additional ablations and analysis of DDC are provided in the \textit{Supplementary Material}.}

\subsection{Inverse Physically-Based Rendering (Inverse PBR)}
\label{PBR}
\jy{COREA provides relightable 3D Gaussians with reliable surface normals via cross-representation geometric coupling, which stabilizes BRDF and lighting decomposition in inverse rendering.
With the learned geometry and normals fixed, we optimize BRDF parameters and lighting in a physically consistent manner.}

\jyn{We supervise per-Gaussian BRDF parameters and environment lighting with an inverse rendering objective, following the formulation of R3DG~\cite{R3DG}.}
To model visibility and shading, we employ the point-based BVH ray tracing scheme from R3DG~\cite{R3DG}, which enables transmittance-aware visibility estimation for Gaussian primitives.
\jyn{We render images by evaluating BRDF responses under the estimated visibility and incident illumination, and supervise them using an $\mathcal{L}_1$ loss against the ground-truth image.}
\jy{Reliable Gaussian normals obtained from the geometric coupling stage improve light-surface interaction modeling, leading to physically faithful rendering and improved relighting quality \jyn{(Tab.~\ref{tab:ablation_normal_comparison}, Fig.~\ref{fig:ablation_normal})}}

\jyn{After optimizing BRDF and lighting, we relight by replacing the global environment map with a novel lighting condition while omitting precomputed indirect illumination, which enables consistent re-evaluation under the new illumination (details in the supplementary material).}
This stage completes COREA’s unified pipeline, \jy{enabling physically-based relighting with high fidelity and consistency.}

%% file: figs/frame_work.tex
\begin{figure*}[!t]
    \centering
    \includegraphics[width=0.98\linewidth]{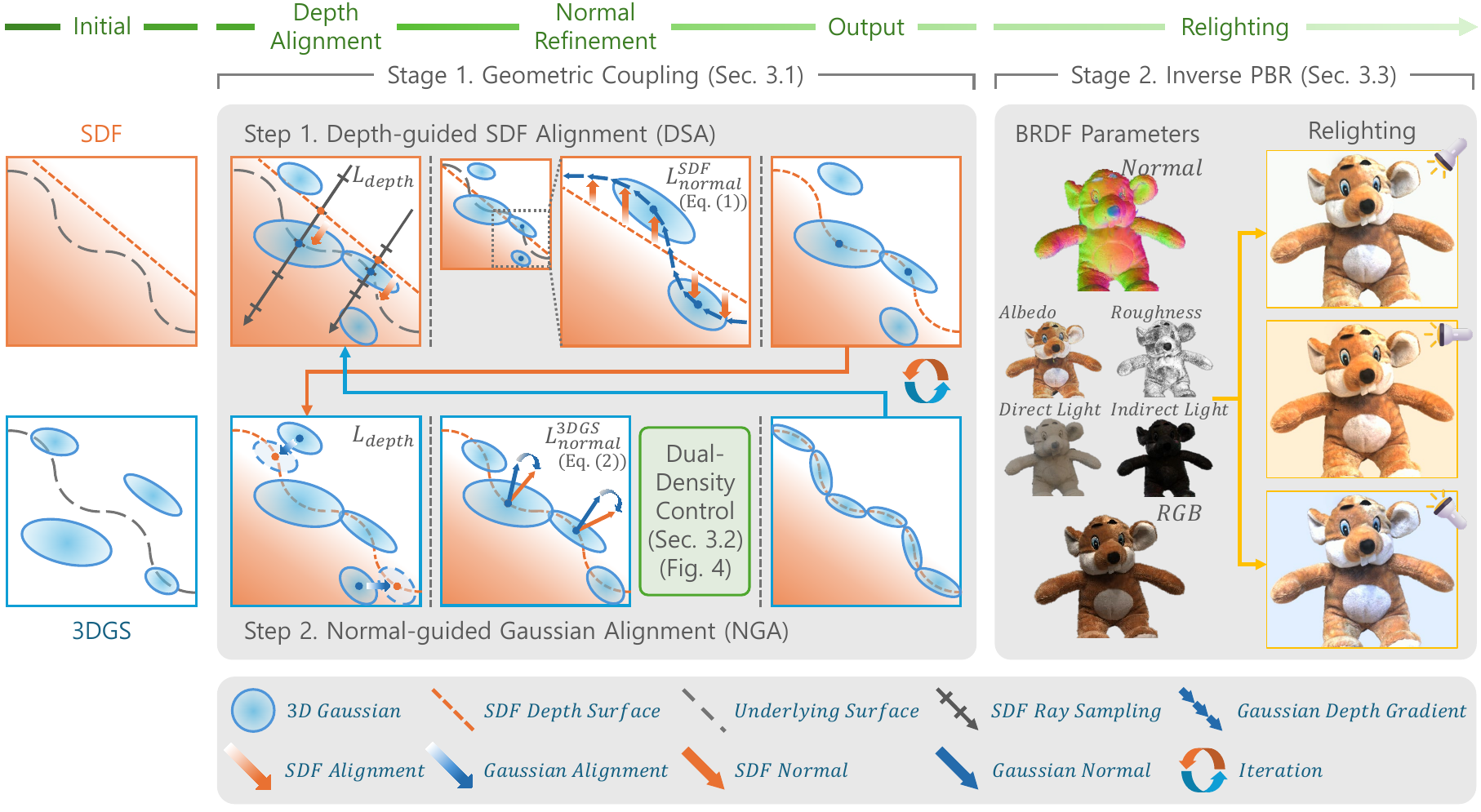}  
    \caption{
    \textbf{Overview of the COREA framework.} 
    \jy{
    COREA jointly trains \jyn{an SDF and relightable 3D Gaussians via geometric coupling on a shared underlying surface, combining reliable depth signals from geometry-constrained relightable 3DGS} with the continuous SDF normal field.
    The coupling uses two complementary modules:
    (i) \textbf{DSA} aligns the SDF to 3DGS via depth-guided ray sampling and matches SDF normals to pixel-wise depth gradients from the 3DGS depth map (Eq.~\eqref{eq:sdf_normal}).
    (ii) \textbf{NGA} guides Gaussians toward the SDF depth surface via depth consistency and aligns Gaussian normals with the continuous SDF normal field, stabilizing normal learning for discrete Gaussians (Eq.~\eqref{eq:3DGS_normal}).
    \textbf{DDC} regulates normal-aware densification using photometric and geometric gradients to suppress redundant splitting.   
    In the second stage, \textbf{Inverse PBR} optimizes BRDF and lighting on the learned geometry, \jyn{enabling accurate BRDF-lighting decomposition and faithful relighting under novel illumination.}}
    }
    \label{fig:framework}
\end{figure*}

%% file: figs/Algorithm.tex
\begin{algorithm}[t]
\caption{\textbf{Training pipeline of COREA}}
\label{alg:corea_train}
\scriptsize
\begin{algorithmic}[1]

\Require Multi-view images $\{\mathbf{I}_m\}_{m=1}^{M}$, cameras $\{\mathbf{C}_m\}_{m=1}^{M}$, SfM point cloud $\mathbf{P}$
\Ensure \BlueNum{SH-based NVS $\widehat{\mathbf{I}}_{\text{NVS}}$}, \OrangeNum{Surface $\mathcal{M}$}, \GreenNum{inverse PBR $\widehat{\mathbf{I}}_{\text{PBR}}$ \&\ relighting $\widehat{\mathbf{I}}_{\text{relight}}$}

\Procedure{TrainCOREA}{$\{\mathbf{I}_m,\mathbf{C}_m\}_{m=1}^{M},\mathbf{P}$}
\State $\mathcal{G}\leftarrow\textsc{InitGaussians}(\mathbf{P})$,\;\;
$\mathcal{F}\leftarrow\textsc{InitSDF}()$

\State \textbf{Warm-up:} Optimize $\mathcal{G}$ w.r.t. $\Lphoto$ for $15\text{k}$

\State \textbf{Stage 1 (Geometric coupling (Sec.~\ref{BDS}, Fig.~\ref{fig:framework})):}
\For{$t=1$ to $30\text{k}$}
    \If{$t \le 10\text{k}$}
        \State $S_{\text{DSA}}=\{\Ldep\}$,\;\; $S_{\text{NGA}}=\{\Ldep\}$
        \Comment{Coarse depth alignment}
    \Else
        \State $S_{\text{DSA}}=\{\Ldep,\LnorS\}$,\;\; $S_{\text{NGA}}=\{\Ldep,\LnorG\}$
        \Comment{Normal refinement}
    \EndIf
    \State \Call{\textbf{Step 1:} $\textsc{DSA}$}{$\mathcal{G},\mathcal{F};\ S_{\text{DSA}}$}\ (\textit{Suppl}.~Alg.~\ref{alg:dsa})
    \Comment{freeze $\mathcal{G}$, update $\mathcal{F}$}
    \State \Call{\textbf{Step 2:} $\textsc{NGA}$}{$\mathcal{G},\mathcal{F};\ S_{\text{NGA}}$}\ (\textit{Suppl}.~Alg.~\ref{alg:nga})
    \Comment{freeze $\mathcal{F}$, update $\mathcal{G}$}

    \If{$t > 10\text{k}$ \textbf{and} $t \,\mathrm{mod}\, 100 = 0$}
        \State \Call{DDC}{$\mathcal{G}$} (Sec.~\ref{DDC}, Fig.~\ref{fig:DDC})
    \EndIf
\EndFor

\State $\mathcal{G}^\star \leftarrow \mathcal{G}$,\quad $\mathcal{F}^\star \leftarrow \mathcal{F}$
\State \textbf{Outputs: }
$\BlueNum{\widehat{\mathbf{I}}_{\text{NVS}}}\leftarrow \textsc{RenderSH}$$(\mathcal{G}^\star,\mathbf{C}_{\text{novel}})$,\;\;
$\OrangeNum{\mathcal{M}}\leftarrow \textsc{MarchingCubes}(\mathcal{F}^\star)$

\State \textbf{Stage 2 (Inverse PBR (Sec.~\ref{PBR}, Fig.~\ref{fig:framework})):}
    \Comment{freeze $\{\mu,\Sigma,\mathbf{n}\}$ of $\mathcal{G}^\star$}
\For{$s=1$ to $10\text{k}$}
    \State Update BRDF and lighting by minimizing $\Lpbr$ (\textit{Suppl}.~\ref{Pre}, \textit{Suppl}.~\ref{Training_objective})
\EndFor

\State \textbf{Outputs: }
$\GreenNum{\widehat{\mathbf{I}}_{\text{PBR}}}\leftarrow \textsc{RenderPBR}(\mathcal{G}^\star,\mathbf{C})$,\;\;
$\GreenNum{\widehat{\mathbf{I}}_{\text{relight}}}\leftarrow \textsc{Relight}(\mathcal{G}^\star,\mathbf{C},l^{env}_{new})$

\EndProcedure
\end{algorithmic}
\end{algorithm}

%% file: figs/DDC.tex
\begin{figure}[!t]  
  \centering\includegraphics[width=0.7\columnwidth]{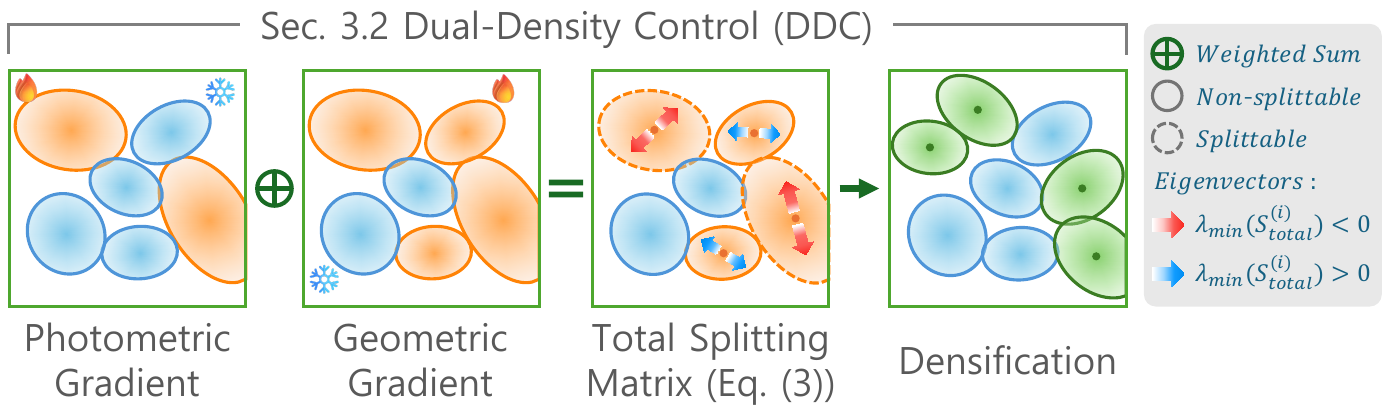}
  \caption{
  \textbf{Dual-Density Control (DDC).}
    Gaussians whose accumulated gradients exceed the densification threshold are marked in red (candidates), while others remain blue.
    For candidates, DDC combines splitting matrices from photometric and geometric gradients into \(S_{\text{total}}\) (Eq.~\eqref{eq:splitting_total}).
    A candidate is split into green Gaussians only if \(\lambda_{\min}(S_{\text{total}}) < 0\), indicating a descent direction that reduces the overall loss.
}
  \label{fig:DDC}
\end{figure}

%% file: sec/4_experiments.tex
\section{Experiments}

\subsection{Experimental Setup}
We evaluate our method on two datasets and three tasks to enable unified assessment of \jyn{SH-based NVS, surface reconstruction, and inverse PBR.}
All experiments are conducted on a single NVIDIA A6000 GPU (48 GB) under the same hardware setting for fair comparison.

\noindent\textbf{Datasets.} 
The DTU dataset~\cite{DTUdataset} provides 15 real-world object scans under controlled lighting and geometry conditions.  
For Tanks\&Temples~\cite{TnTdataset}, we follow the R3DG~\cite{R3DG} setup and use the four object-centric scenes defined in their work, \jyn{where evaluation focuses on foreground objects.}
This setup enables consistent evaluation of geometry and appearance under real-world conditions.

\noindent\textbf{Evaluation Metrics.} 
For \jyn{SH-based NVS}, we compare all methods using PSNR, SSIM~\cite{SSIM}, and LPIPS~\cite{LPIPS}. 
For \jyn{surface reconstruction}, we evaluate methods that output meshes and measure geometric accuracy using Chamfer Distance (CD).
\jyn{Inverse PBR} evaluation is conducted only for models capable of BRDF-based rendering, using the same metrics as \jyn{SH-based NVS.}

\input{tab/overall_qualitative_combine}

\subsection{Evaluation Results}
We render qualitative results on a white background for consistent visualization across tasks, emphasizing overall rendering quality and clear separation between objects and the background.
Black speckles or dark edge artifacts in some methods indicate opacity noise often caused by excessive Gaussian density.
Methods that do not support specific tasks are marked as \textit{N/S}, and those that exceed the available GPU memory during training under the same single-GPU setting are marked as \textit{OOM}.
COREA unifies an SDF and relightable 3DGS through geometric coupling on a shared underlying surface, and is the only framework that supports \jyn{SH-based NVS}, \jyn{surface reconstruction}, and \jyn{inverse PBR} within a single pipeline.
COREA achieves \jyn{competitive SH-based NVS} and \jyn{surface reconstruction} performance, while delivering particularly strong \jyn{inverse PBR} results.
Additional per-scene results are provided in the \textit{Supplementary Material} and \href{https://cau-vilab.github.io/COREA/}{\textit{demo video}}.

\subsubsection{SH-based Novel-View Synthesis Comparison.}
We evaluate SH-based NVS performance on the DTU~\cite{DTUdataset} and Tanks\&Temples~\cite{TnTdataset} datasets.
As summarized in Tab.~\ref{tab:combined} (SH), 
Our method achieves strong performance on DTU across PSNR, SSIM, and LPIPS, demonstrating robust rendering on real-world captures.
On Tanks\&Temples, the performance is slightly below the top baseline, \jyn{likely due to the larger scene scale and sparser camera coverage, which makes joint optimization and cross-representation alignment more challenging.}
Nevertheless, our framework achieves strong \jyn{inverse PBR performance} (Sec.~\ref{sec:pbr}) and \jyn{competitive surface reconstruction quality} (Sec.~\ref{sec:mesh_recon}) on this dataset, while being the only method that supports all three tasks within a \textit{unified} framework.

As shown in Fig.~\ref{fig:main_comparison} (blue boxes), our method produces sharper and more coherent novel views with fewer background artifacts than other approaches.
These results \jyn{reflect stable geometric coupling and precise cross-representation alignment between the SDF and relightable 3DGS, which are essential for stable relighting under environment-map replacement.}

\input{figs/main_comparison}

\subsubsection{Surface Reconstruction Comparison.}
\label{sec:mesh_recon}
\jyn{We further evaluate surface reconstruction to assess the complementary geometric interaction enabled by coupling an SDF and geometry-constrained relightable 3DGS, and compare COREA with explicit-implicit hybrid baselines~\cite{GSDF, GS_pull, GaussianOpacityFields, GSROR2} on DTU using both quantitative and qualitative results.
As summarized in Tab.~\ref{tab:combined}~(CD), COREA demonstrates competitive CD performance compared to existing explicit-implicit hybrid baselines, indicating reliable surface reconstruction.
Qualitative results in Fig.~\ref{fig:main_comparison} (orange boxes) further demonstrate that COREA reconstructs sharper and more detailed surfaces, preserving fine-scale geometric structures that are often oversmoothed by baselines.}

This improvement stems from our geometric coupling between an SDF and geometry-constrained relightable 3DGS on a shared underlying surface, where pixel-wise depth gradients from the 3DGS depth map provide a strong signal for refining the SDF geometry.
The resulting surfaces also provide reliable normals for inverse PBR optimization.

\input{figs/relight_comparision}

\subsubsection{Inverse Physically-Based Rendering Comparison.}
\label{sec:pbr}
We evaluate our framework for \jyn{inverse PBR} against prior relightable Gaussian approaches~\cite{GaussianShader, R3DG, GSIR}.
As shown in Tab.~\ref{tab:combined} ({PBR}), our method achieves the highest PBR quality among relightable baselines, indicating that reliable geometry and normals are critical for accurate shading and reflectance recovery.

For qualitative evaluation, Fig.~\ref{fig:main_comparison} (green boxes) presents PBR renderings under the original illumination.
Our method reproduces fine-scale surface details more faithfully, as seen in the normal patches at the bottom right, where surface orientations are accurately captured and reflected in the rendered appearance.

Extended results are provided in Fig.~\ref{fig:relight_comparison}. 
The first row shows the same PBR reconstruction, and the subsequent rows show relighting under novel illumination with changed light directions and environment maps.
COREA maintains consistent shading and reflectance across lighting changes, producing clean foreground silhouettes with reduced opacity noise and fewer boundary artifacts.
This robustness stems from our geometric coupling between an SDF and relightable 3DGS on a shared underlying surface, which stabilizes normal learning and supports reliable relighting under viewpoint changes.

%% file: tab/overall_qualitative_combine.tex
\begin{table}[t]
\caption{
\textbf{Quantitative Comparison.} 
We evaluate SH-based NVS, surface reconstruction, and inverse PBR on the DTU and Tanks\&Temples datasets.
All methods are evaluated under the same single-GPU setting for a fair comparison.
\textit{N/S} denotes methods that do not support the corresponding task, and \textit{OOM} indicates failures under the same GPU memory budget.
COREA achieves strong performance in SH-based NVS and inverse PBR with competitive surface reconstruction (CD), while uniquely supporting all three tasks within a unified framework.
The best, second-best, and third-best results are highlighted in red, yellow, and purple, respectively.
}
\centering
\renewcommand{\arraystretch}{1.1}
\setlength{\tabcolsep}{4.5pt}

\resizebox{\textwidth}{!}{%

\begin{tabular}{l|l|cccc|ccc||ccc|ccc}
  \toprule
  & \textbf{Dataset} &
  \multicolumn{4}{c|}{\textbf{DTU (SH)}} &
  \multicolumn{3}{c||}{\textbf{Tanks\&Temples (SH)}} &
  \multicolumn{3}{c|}{\textbf{DTU (PBR)}} &
  \multicolumn{3}{c}{\textbf{Tanks\&Temples (PBR)}} \\ 
  \cmidrule(lr){3-6}\cmidrule(lr){7-9}\cmidrule(lr){10-12}\cmidrule(lr){13-15}
  \textbf{Category} & \textbf{Method} &
  CD$\downarrow$ & PSNR$\uparrow$ & SSIM$\uparrow$ & LPIPS$\downarrow$ &
  PSNR$\uparrow$ & SSIM$\uparrow$ & LPIPS$\downarrow$ &
  PSNR$\uparrow$ & SSIM$\uparrow$ & LPIPS$\downarrow$ &
  PSNR$\uparrow$ & SSIM$\uparrow$ & LPIPS$\downarrow$ \\
  \midrule
  Vanilla &
  3DGS (SIGGRAPH’23) & 
  \textit{N/S} & 27.30 & 0.867 & 0.184 &
  27.46 & 0.910 & 0.115 &
  \textit{N/S} & \textit{N/S} & \textit{N/S} &
  \textit{N/S} & \textit{N/S} & \textit{N/S} \\
  \midrule
  Relitable &
  GaussianShader (CVPR’24) &
  \textit{N/S} & 22.14 & 0.826 & 0.201 &
  19.96 & 0.851 & 0.139 &
  19.49 & 0.719 & 0.393 &
  17.82 & \cellcolor{third}0.829 & \cellcolor{third}0.169 \\
  &
  GS-IR (CVPR’24) & 
  \textit{N/S} & 28.29 & 0.874 & 0.173 &
  \cellcolor{third}28.93 & 0.929 & \cellcolor{third}0.089 &
  \cellcolor{third}24.56 & \cellcolor{third}0.752 & \cellcolor{third}0.206 &
  \cellcolor{third}22.96 & 0.652 & 0.186 \\
  &
  R3DG (ECCV’24) & 
  \textit{N/S} & \cellcolor{second}31.82 & \cellcolor{second}0.944 & \cellcolor{third}0.105 &
  \cellcolor{best}29.92 & \cellcolor{best}0.941 & \cellcolor{second}0.079 &
  \cellcolor{second}26.44 & \cellcolor{second}0.914 & \cellcolor{second}0.138 &
  \cellcolor{second}26.31 & \cellcolor{second}0.912 & \cellcolor{second}0.103 \\
  &
  SVG-IR (CVPR’25) & 
  \textit{N/S} & 28.02 & \cellcolor{third}0.923 & \cellcolor{best}0.055 &
  27.68 & \cellcolor{third}0.934 & \cellcolor{best}0.045 &
  \textit{OOM} & \textit{OOM} & \textit{OOM} &
  \textit{OOM} & \textit{OOM} & \textit{OOM} \\
  \midrule
  Hybrid &
  GOF (SIGGRAPH’24) & 
  \cellcolor{best}0.802 & \cellcolor{third}28.70 & 0.909 & 0.124 &
  \textit{OOM} & \textit{OOM} & \textit{OOM} &
  \textit{N/S} & \textit{N/S} & \textit{N/S} &
  \textit{N/S} & \textit{N/S} & \textit{N/S} \\
  &
  GSDF (NeurIPS’24) & 
  \cellcolor{third}0.859 & 27.76 & 0.874 & 0.176 &
  \textit{OOM} & \textit{OOM} & \textit{OOM} &
  \textit{N/S} & \textit{N/S} & \textit{N/S} &
  \textit{N/S} & \textit{N/S} & \textit{N/S} \\
  &
  GS-Pull (NeurIPS’24) & 
  0.916 & 27.36 & 0.872 & 0.208 &
  26.29 & 0.893 & 0.142 &
  \textit{N/S} & \textit{N/S} & \textit{N/S} &
  \textit{N/S} & \textit{N/S} & \textit{N/S} \\
  &
  GS-ROR (TOG’25) & 
  2.552 & \textit{N/S} & \textit{N/S} & \textit{N/S} &
  \textit{N/S} & \textit{N/S} & \textit{N/S} &
  20.38 & 0.850 & 0.206 &
  22.44 & 0.887 & 0.145 \\
  &
  \textbf{COREA (Ours)} & 
  \cellcolor{second}0.824 & \cellcolor{best}32.27 & \cellcolor{best}0.955 & \cellcolor{second}0.094 &
  \cellcolor{second}29.25 & \cellcolor{second}0.937 & \cellcolor{third}0.089 &
  \cellcolor{best}29.72 & \cellcolor{best}0.942 & \cellcolor{best}0.104 &
  \cellcolor{best}27.57 & \cellcolor{best}0.927 & \cellcolor{best}0.102 \\
  \bottomrule
\end{tabular}}
\label{tab:combined}
\end{table}

%% file: figs/main_comparison.tex
\begin{figure}[t]
    \centering
    \includegraphics[width=0.98\linewidth]{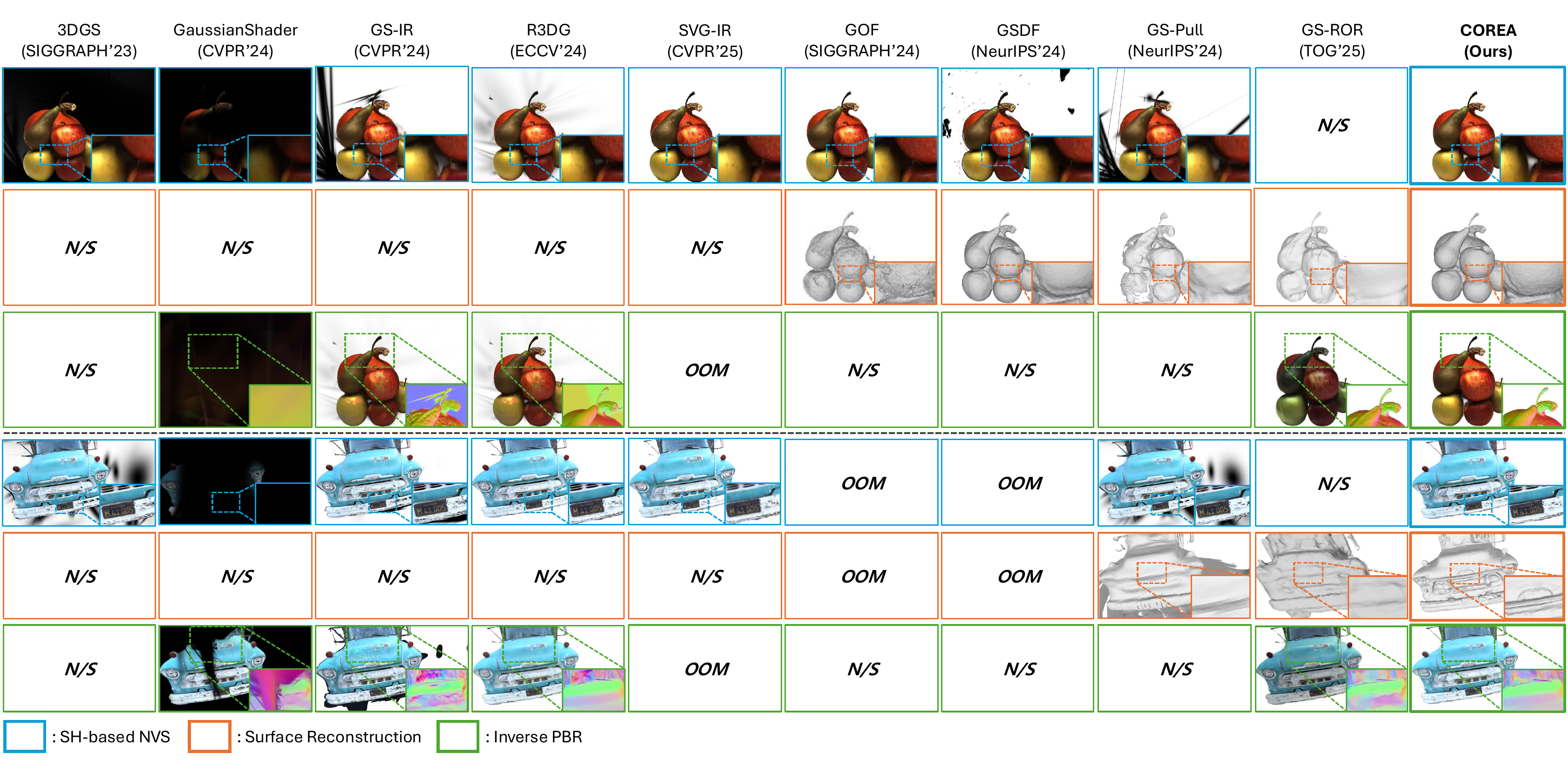}
    \caption{
    \textbf{Qualitative Comparison.}
    We compare \textbf{COREA} with recent Gaussian-based methods on three tasks: \jyn{SH-based NVS} (blue), \jyn{surface reconstruction} (orange), and \jyn{inverse PBR} (green).
    All results are rendered on a white background for consistent visual comparison.
    Artifacts such as black background patches or dark speckles observed in some baselines stem from excessive Gaussian opacity; by contrast, \textbf{COREA} produces clean, artifact-free renderings on white backgrounds.
    Our method is \jyn{the only framework that supports} all three tasks under the same GPU setting, whereas others show \textit{N/S} (Not Supported) or \textit{OOM} (Out of Memory).
    Overall, COREA yields sharper novel views, more faithful BRDF and lighting decomposition, and finer geometric details through complementary cross-representation geometric coupling.
    Additional qualitative results are provided in the \textit{Supplementary Material} and \href{https://cau-vilab.github.io/COREA/}{\textit{demo video}}.
    }
    \label{fig:main_comparison}
\end{figure}

%% file: figs/relight_comparision.tex
\begin{figure}[!tbp]
    \centering
    \includegraphics[width=0.98\textwidth]{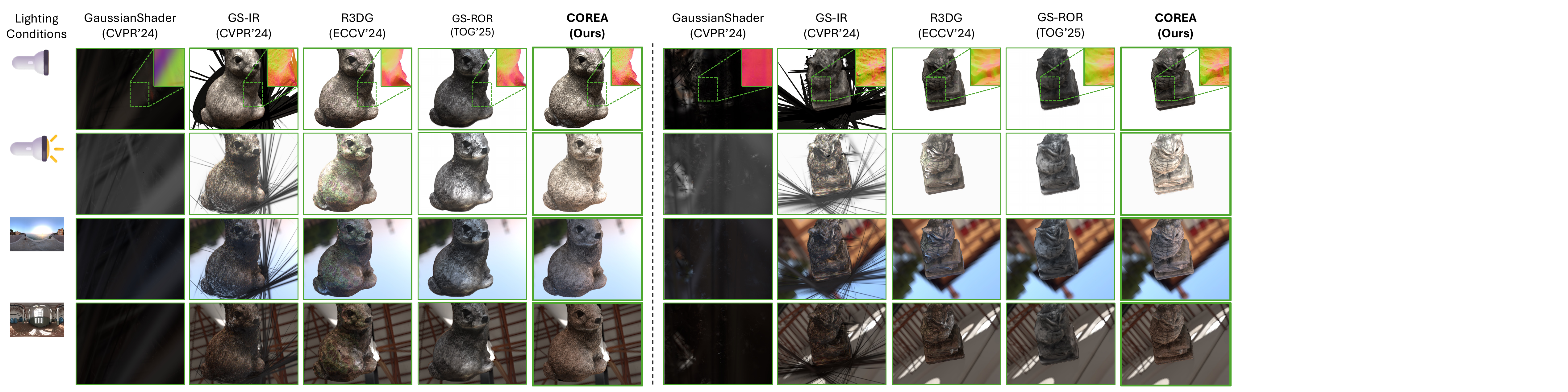}
    \caption{
    \textbf{Qualitative Results of Inverse PBR and Relighting.}
    We compare \textbf{COREA} with recent relightable Gaussian-based methods under varying illumination conditions.
    The first row shows PBR renderings under the original lighting setup, while the remaining rows illustrate relighting results under directional lights and various HDR environment maps.
    Compared to previous methods that exhibit black background artifacts or unstable reflectance under varying illumination, \textbf{COREA} maintains consistent shading, clean appearance, and accurate reflectance reconstruction across all lighting conditions, demonstrating robust BRDF and geometry alignment.
    }
    \label{fig:relight_comparison}
\end{figure}

%% file: sec/5_ablation.tex
\subsection{Ablation Study}

\subsubsection{Depth-Gradient Supervision for Geometry Alignment.}
\label{ablation_normal_comparison}
\begin{figure}[!t]
\centering

\begin{minipage}{0.49\linewidth}
\centering
\captionsetup{type=table}
\caption{
\textbf{Quantitative Comparison of Normal Supervision Strategies.}
We compare alpha-blended normal supervision and our pixel-wise depth gradient supervision for SDF alignment.
Pixel-wise depth gradients provide more accurate geometric supervision, enabling more stable convergence of the SDF alignment and improving surface reconstruction (CD), SH-based NVS and inverse PBR performance.
}
\renewcommand{\arraystretch}{1.1}
\resizebox{\linewidth}{!}{%
\begin{tabular}{l|c|ccc|ccc}
  \toprule
  & \multicolumn{1}{c|}{\textbf{Mesh}} & \multicolumn{3}{c|}{\textbf{SH-based NVS}} & \multicolumn{3}{c}{\textbf{Inverse PBR}} \\ 
  \cmidrule(lr){2-2}\cmidrule(lr){3-5}\cmidrule(lr){6-8}
  \textbf{Method} & \textbf{CD$\downarrow$} &
  \textbf{PSNR$\uparrow$} & \textbf{SSIM$\uparrow$} & \textbf{LPIPS$\downarrow$} &
  \textbf{PSNR$\uparrow$} & \textbf{SSIM$\uparrow$} & \textbf{LPIPS$\downarrow$} \\
  \midrule
  Alpha-blended normal & 0.848 & 31.89 & 0.947 & 0.096 & 29.38 & 0.933 & 0.116 \\
  Pixel-wise depth gradient (Ours) & \textbf{0.824} & \textbf{32.27} & \textbf{0.955} & \textbf{0.094} &
                       \textbf{29.72} & \textbf{0.942} & \textbf{0.104} \\
  \bottomrule
\end{tabular}
}
\label{tab:ablation_normal_comparison}
\end{minipage}
\hfill
\begin{minipage}{0.49\linewidth}
\centering
\includegraphics[width=\linewidth]{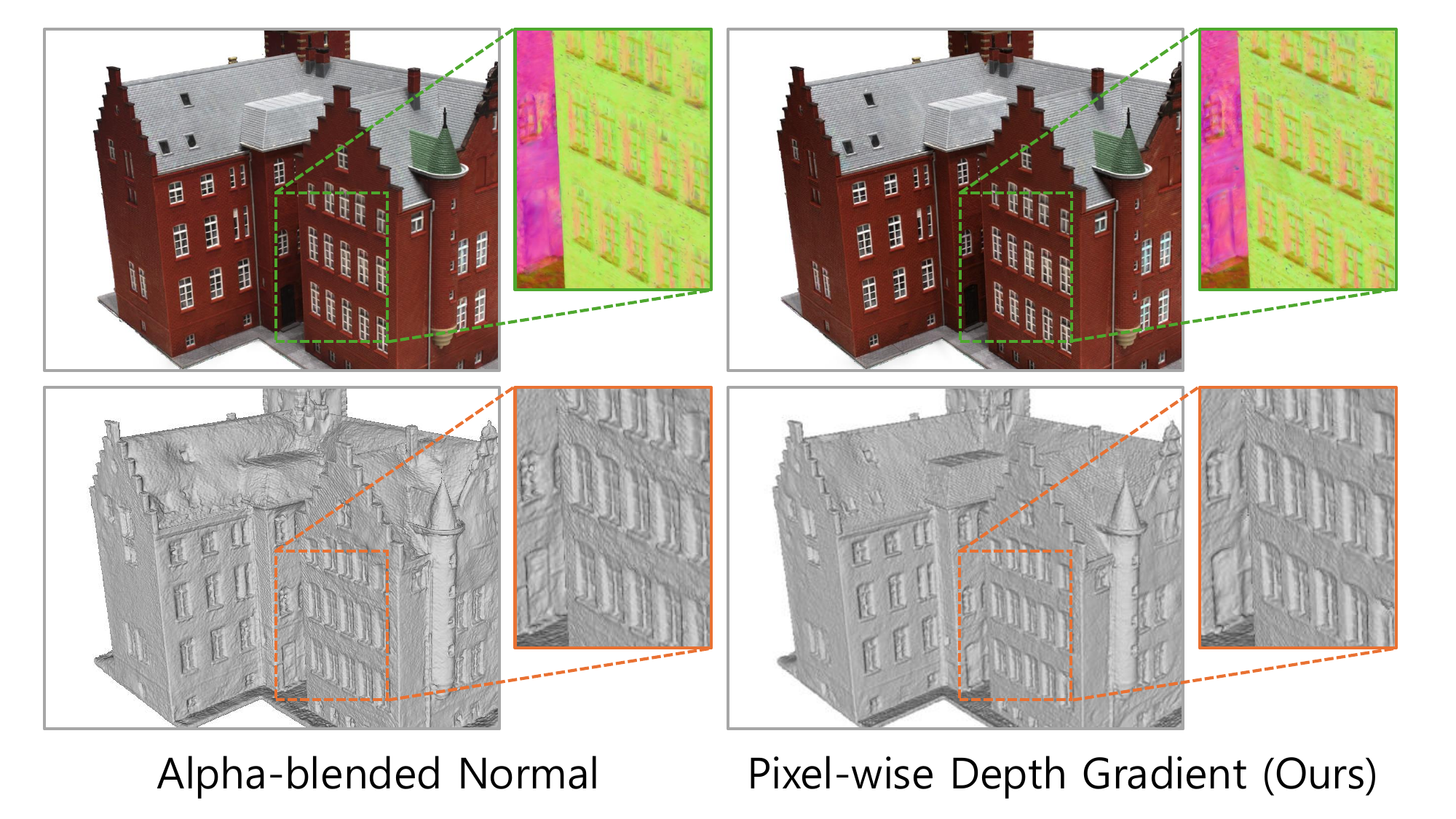}
\vspace{-10pt}
\captionsetup{type=figure}
\caption{
\textbf{Qualitative Comparison of Normal Supervision Strategies.}
Upper row: inverse PBR renderings. Lower row: reconstructed surfaces.
Pixel-wise depth gradient supervision yields more faithful relighting and more accurate surface reconstruction.
}
\label{fig:ablation_normal}
\end{minipage}
\end{figure}

\jyn{We conduct an ablation study to validate pixel-wise depth gradient supervision for geometry alignment in our coupled SDF and relightable 3DGS framework.
Fig.~\ref{fig:ablation_normal} qualitatively compares alpha-blended normal supervision and depth-gradient supervision.
The upper row shows inverse-PBR renderings with normal patches, while the lower row shows reconstructed surfaces with zoomed-in regions.
With alpha-blended normal supervision, the normal patches exhibit weaker shading contrast and the reconstructed surfaces appear inflated with missing fine-scale structure.
This degradation arises because the discrete Gaussians oversmooth rendered normals through alpha blending, weakening SDF supervision and allowing errors to propagate back to 3DGS through the coupled optimization.
In contrast, pixel-wise depth gradient supervision preserves sharper local structure and yields more refined surface geometry and normals.}

\jyn{Tab.~\ref{tab:ablation_normal_comparison} provides quantitative results across \jyn{SH-based NVS}, \jyn{surface reconstruction}, and \jyn{inverse PBR}.
Compared to alpha-blended normal supervision, pixel-wise depth gradient supervision consistently improves rendering quality and geometric accuracy, achieving lower Chamfer Distance (CD) and higher PSNR and SSIM with lower LPIPS for \jyn{SH-based NVS} and \jyn{inverse PBR}.
Overall, depth-gradient supervision provides a stronger geometric signal than alpha-blended normals, leading to more reliable geometric coupling and improved fidelity across tasks.}

\subsubsection{Geometric Coupling for Reliable Alignment.}
\jyn{We analyze the role of each alignment module by disabling it while keeping the rest of the pipeline unchanged; quantitative results are summarized in Tab.~\ref{tab:ablation_alignment}.}

\noindent\textbf{Module ablations.}
\jyn{Each ablation breaks one direction of our geometric coupling loop.
In \textbf{w/o DSA}, the SDF no longer receives depth-based guidance from geometry-constrained relightable 3DGS (depth alignment and depth-gradient normal refinement), leading to a clear drop in surface reconstruction quality, particularly in CD.
In \textbf{w/o NGA}, relightable 3DGS no longer benefits from the continuous SDF normal field, which prevents stable Gaussian normal formation and weakens normal-aware densification, causing the largest degradation in inverse PBR where sharp and consistent normals are essential.}

Together, these results confirm that COREA derives its strength from coupling complementary geometric properties on a shared underlying surface: DSA anchors and refines the SDF using \jyn{geometry-constrained depth signals} from relightable 3DGS, while NGA stabilizes Gaussian normals using the continuous SDF normal field, yielding reliable geometry and improved inverse PBR quality.

\noindent\textbf{Update strategy.}
\jyn{We also compare two update strategies for the full pipeline (Alg.~\ref{alg:corea_train}): a simultaneous scheme that updates DSA and NGA in the same step (Full-Simul.) and our alternative scheme that updates them in alternating steps (Full-Alt.).
As shown in Tab.~\ref{tab:ablation_alignment}, the alternative strategy performs best overall, as alternating updates reduce interference and yield more stable optimization by updating one representation while keeping the other fixed. 
We adopt the alternative strategy by default.}

\input{tab/ablation_alignment}

\vspace{-5pt}

%% file: tab/ablation_alignment.tex
\begin{table}[!t]
\caption{
\textbf{Ablation of Alignment Modules and Update Strategies.}
\textbf{w/o DSA} removes depth-based guidance from geometry-constrained relightable 3DGS for SDF alignment and depth-gradient refinement, leading to less precise SDF geometry and a clear degradation in surface reconstruction (CD).
\textbf{w/o NGA} removes SDF-guided depth alignment and continuity-based normal refinement for relightable 3DGS, yielding oversmoothed Gaussian normals and the largest drop in inverse PBR.
With both modules enabled, \textbf{Full-Alt.} achieves the best overall performance by reducing interference through alternating updates, outperforming \textbf{Full-Simul.} on SH-based NVS, surface reconstruction, and inverse PBR.}
\centering
\renewcommand{\arraystretch}{1.1}
\resizebox{0.6\textwidth}{!}{%
\begin{tabular}{l|c|ccc|ccc}
  \toprule
  & \multicolumn{1}{c|}{\textbf{Mesh}} & \multicolumn{3}{c|}{\textbf{NVS}} & \multicolumn{3}{c}{\textbf{PBR}} \\ 
  \cmidrule(lr){2-2}\cmidrule(lr){3-5}\cmidrule(lr){6-8}
  \textbf{Setting} & \textbf{CD$\downarrow$} &
  \textbf{PSNR$\uparrow$} & \textbf{SSIM$\uparrow$} & \textbf{LPIPS$\downarrow$} &
  \textbf{PSNR$\uparrow$} & \textbf{SSIM$\uparrow$} & \textbf{LPIPS$\downarrow$} \\
  \midrule
  \textit{w/o} NGA & 0.863 & 31.47 & 0.944 & 0.104 & 28.63 & 0.926 & 0.124 \\
  \textit{w/o} DSA & 0.932 & 31.50 & 0.945 & 0.097 & 28.23 & 0.908 & 0.132 \\
  Full-Simul.  & 0.852 & 31.62 & 0.946 & 0.105 & 29.07 & 0.931 & 0.123 \\
  Full-Alt. (COREA) & \textbf{0.824} & \textbf{32.27} & \textbf{0.955} & \textbf{0.094} &
                  \textbf{29.72} & \textbf{0.942} & \textbf{0.104} \\
  \bottomrule
\end{tabular}
}
\label{tab:ablation_alignment}
\end{table}

%% file: sec/6_conclusion.tex
\section{Conclusion}
We presented COREA, the first unified \jyn{three-tasks framework that jointly learns an SDF and relightable 3D Gaussians by coupling their complementary geometric properties on a shared underlying surface.}
\jyn{By anchoring SDF geometry with depth signals and pixel-wise depth gradients from geometry-constrained relightable 3DGS and stabilizing Gaussian normal learning using the continuous SDF normal field, COREA preserves fine-scale structure and provides reliable normals.
We further introduce Dual-Density Control, which regularizes normal-aware densification using both photometric and geometric gradients to prevent excessive Gaussian growth while maintaining stable rendering.}
These components provide a consistent geometric foundation for \jyn{inverse PBR, enabling accurate BRDF and lighting decomposition and faithful relighting under novel illumination.}
Experiments show that \jyn{COREA achieves strong results in SH-based NVS and inverse PBR while delivering competitive surface reconstruction, and remains the only framework that supports all three tasks.}

%% file: sec/X_suppl.tex
\section*{Supplementary Material}
\addcontentsline{toc}{section}{Supplementary Material}

\setcounter{figure}{0}
\setcounter{table}{0}
\setcounter{equation}{0}
\renewcommand{\thefigure}{S\arabic{figure}}
\renewcommand{\thetable}{S\arabic{table}}
\renewcommand{\theequation}{S\arabic{equation}}




\section{Demo Video} 
To present the proposed method more effectively, we provide a \href{https://cau-vilab.github.io/COREA/}{\textit{demo video}}. 
The video is organized into several segments, including an overall teaser, a brief visualization of our framework, and additional visual materials demonstrating the qualitative performance of COREA. 
It also presents comparisons across three key tasks: SH-based NVS, surface reconstruction, and inverse PBR.
The final demo is edited using Adobe After Effects~\cite{adobe_ae}. 
Source clips for SH-based NVS and inverse PBR are generated through the web-based renderer Supersplat~\cite{supersplat}, and clips for surface reconstruction are rendered using Blender~\cite{blender}.

\section{Implementation details}

\subsection{Inverse Rendering Background}
\label{Pre}
\textbf{Inverse rendering with 3DGS.}
Following the inverse rendering formulation of R3DG~\cite{R3DG}, each Gaussian primitive is augmented with BRDF parameters, including albedo $b$ and roughness $r$, together with a learned indirect illumination term $l$ and a shared global environment map $l^{env}$.
The diffuse component $f_d$ is determined by $b$, while the specular component $f_s$ depends on the roughness $r$ as well as the incoming and outgoing directions $\omega_i$ and $\omega_o$, together defining the BRDF response of each Gaussian.

\noindent\textbf{Incident radiance and physically-based color.}
Given a visibility term $V(\omega_i)$ along direction $\omega_i$, the incident radiance is modeled as the sum of the visible environment lighting and a learned indirect illumination term:
{
\begin{align}
L(\omega_i) &= V(\omega_i)\, l^{env}(\omega_i) + l(\omega_i),
\label{eq:incident_radiance} \\
c'(\omega_o) &=
\sum_{i=1}^{N_s} \left[f_d + f_s(\omega_o,\omega_i)\right]
L(\omega_i)\,(\omega_i \cdot \mathbf{n})\,\Delta\omega_i,
\label{eq:pbr_color}
\end{align}
}
where $L_i(\omega_i)$ denotes the incident radiance composed of the environment light $l^{env}$ and the learned indirect term $l_i$, 
$\mathbf{n}$ is the Gaussian normal,
$c'(\omega_o)$ is the outgoing radiance toward $\omega_o$,
$N_s$ is the number of sampled incident directions,
and $\Delta\omega_i$ is the solid-angle weight of the $i$-th sample.
The physically-based color $c'(\omega_o)$ is rendered by evaluating Eq.~\eqref{eq:pbr_color} under the incident radiance defined in Eq.~\eqref{eq:incident_radiance}.
For relighting under a novel illumination, we replace the global environment map $l^{env}$ with $l^{env}_{\text{new}}$ and re-evaluate Eq.~\eqref{eq:pbr_color} using the updated incident radiance, while omitting the learned indirect illumination term $l(\omega_i)$ for consistent rendering under the new lighting.

This formulation enables spatially varying relighting while remaining compatible with the standard SH-based rendering pipeline of 3DGS.
We estimate visibility $V(\omega_i)$ following the BVH-based ray tracing scheme of R3DG~\cite{R3DG}, which models Gaussian transmittance along sampled incident rays.

\subsection{Training Objectives}
\label{Training_objective}
We adopt a two-stage training scheme: geometric coupling in the first stage, followed by inverse PBR in the second stage.

In the geometric coupling stage, an SDF and a relightable 3DGS are jointly optimized through depth and normal alignment with alternating updates for stable training.
Accordingly, the total geometry loss is:
\begin{equation}
\mathcal{L}_{\text{Geometry}} =
\mathcal{L}_{\text{SDF}} + \mathcal{L}_{\text{3DGS}}.
\end{equation}

We first define a cross-representation depth consistency loss, which align the depth maps rendered from the SDF and the geometry-constrained relightable 3DGS using an $\mathcal{L}_1$ loss.
This coarse alignment encourages the two representations to converge to a shared geometric structure:
\begin{equation}
\mathcal{L}_{\text{depth}} = \frac{1}{N} \sum_{i=1}^{N} 
\left| \mathbf{d}_{\text{SDF}}^{(i)} - \mathbf{d}_{\text{3DGS}}^{(i)} \right|
\label{eq:supp_Ldep}
\end{equation}
where $\mathbf{d}_{\text{SDF}}$ and $\mathbf{d}_{\text{3DGS}}$ denote the rendered depth maps of the SDF and relightable 3DGS, respectively, and $N$ is the number of valid pixels. 

\noindent The loss function for the SDF network is:
\begin{multline}
\mathcal{L}_{\text{SDF}} =
\lambda_{1}^{\text{SDF}}   \mathcal{L}_{1}^{\text{SDF}} +
\lambda_{\text{depth}} \mathcal{L}_{\text{depth}} +
\lambda_{\text{eik}}   \mathcal{L}_{\text{eik}} + \\
\lambda_{\text{curv}}  \mathcal{L}_{\text{curv}} +
\lambda_{\text{normal}} \mathcal{L}_{\text{normal}}^{\text{SDF}}
\label{eq:supp_LSDF}
\end{multline}
where $\mathcal{L}_{1}^{\text{SDF}}$ is a reconstruction term used for coarse alignment,
$\mathcal{L}_{\text{eik}}$ enforces the Eikonal constraint~\cite{EikonalReg},
$\mathcal{L}_{\text{curv}}$ encourages geometric smoothness~\cite{neuralangelo, permutosdf},
and $\mathcal{L}_{\text{normal}}^{\text{SDF}}$ (Eq.~\eqref{eq:sdf_normal}) aligns the SDF gradient with the pixel-wise depth gradient from 3DGS, as described in DSA (Sec.~\ref{DSA}).

\noindent The loss function for the 3DGS network is defined as:
\begin{multline}
\mathcal{L}_{\text{3DGS}} =
\mathcal{L}_{\text{image}}^{\text{3DGS}}+
\lambda_{\text{depth}} \mathcal{L}_{\text{depth}} + 
\lambda_O \mathcal{L}_O +
\lambda_u \mathcal{L}_u +
\lambda_{\text{normal}} \mathcal{L}_{\text{normal}}^{\text{3DGS}}
\label{eq:supp_L3DGS}
\end{multline}
Here, $\mathcal{L}_O$ and $\mathcal{L}_u$ are the mask and depth-distribution constraints adopted from R3DG~\cite{R3DG}.
In particular, $\mathcal{L}_u$ regularizes the rendered depth distribution by minimizing its uncertainty,
e.g., $\mathcal{L}_u = \mathbb{E}[d^2] - (\mathbb{E}[d])^2$ with $\mathbb{E}[d]=\sum_{i\in \mathcal{K}} w_i d_i$ and $\mathbb{E}[d^2]=\sum_{i\in \mathcal{K}} w_i d_i^2$, where $w_i$ denotes the normalized rendering weight defined in R3DG~\cite{R3DG}.
These constraints suppress noise in the rendered depth map $\mathbf{d}_{\text{3DGS}}$, and consequently stabilize the pixel-wise depth gradient $\mathbf{n}_{\text{3DGS}}^{\text{DG}}=\nabla \mathbf{d}_{\text{3DGS}}/\|\nabla \mathbf{d}_{\text{3DGS}}\|$.
As a result, $\mathbf{d}_{\text{3DGS}}$ and $\mathbf{n}_{\text{3DGS}}^{\text{DG}}$ provide sharper and more reliable supervision signals than alpha-blended normals (Fig.~\ref{fig:normal_comparison}), making them more stable supervision signals for DSA.

\noindent We define the photometric objective used in densification as
\begin{equation}
\mathcal{L}_{\text{image}}^{\text{3DGS}} =
\lambda_{1}^{\text{3DGS}} \mathcal{L}_1^{\text{3DGS}} +
\lambda_{\text{ssim}} \mathcal{L}_{\text{ssim}},
\label{eq:Limage_def}
\end{equation}
whose gradients correspond to the photometric gradients used in DDC.
Similarly, $\mathcal{L}_{\text{normal}}^{\text{3DGS}}$ (Eq.~\eqref{eq:3DGS_normal}) is the cosine loss used in NGA to align 3DGS normals with SDF normals, and its gradients constitute the geometric gradients used in DDC for normal-aware densification. (Sec.~\ref{DDC})

After geometric coupling, we optimize BRDF parameters and lighting via inverse PBR.
Following R3DG~\cite{R3DG}, we sample $N_s = 64$ incident directions for each Gaussian and optimize the rendered appearance through physically-based rendering supervision against the ground-truth image.
The inverse PBR loss is defined as:
\begin{equation}
\mathcal{L}_{\text{PBR}} =
\lambda_1 \mathcal{L}_1 +
\lambda_{\text{ssim}} \mathcal{L}_{\text{ssim}} +
\lambda_{\text{light}} \mathcal{L}_{\text{light}} +
\lambda_{\text{BRDF}} \mathcal{L}_{\text{BRDF}}.
\label{PBRloss}
\end{equation}
Here, $\mathcal{L}_{\text{light}}$regularizes lighting estimation for stable decomposition of illumination, while
$\mathcal{L}_{\text{BRDF}}$ regularizes the BRDF parameters.

\begin{algorithm}[t]
\caption{\textbf{Step 1: Depth-guided SDF Alignment (DSA)}}
\label{alg:dsa}
\small
\begin{algorithmic}[1]
\Procedure{DSA}{$\mathcal{G},\mathcal{F};\ S_{\text{DSA}}$}
\State \textbf{Freeze} $\mathcal{G}$

\State $\mathbf{d}_{\text{3DGS}} \leftarrow \textsc{RenderDepth}(\mathcal{G})$
\State $(\mathbf{d}_{\text{SDF}},\mathbf{n}_{\text{SDF}}) \leftarrow \textsc{RenderSDF}(\mathcal{F})$

\If{$\Ldep \in S_{\text{DSA}}$}
    \State Compute $\Ldep(\mathbf{d}_{\text{SDF}},\mathbf{d}_{\text{3DGS}})$
    \Comment{\textit{Suppl}.~Sec.~\ref{Training_objective}, Eq.~\eqref{eq:supp_Ldep}}
\EndIf
\If{$\LnorS \in S_{\text{DSA}}$}
    \State $\widetilde{\mathbf{n}}_{\text{3DGS}} \leftarrow \textsc{Normalize}(\nabla \mathbf{d}_{\text{3DGS}})$ 
    \Comment{\textit{Main}.~Fig.~\ref{fig:normal_comparison}~(B)}
    \State Compute $\LnorS(\mathbf{n}_{\text{SDF}},\widetilde{\mathbf{n}}_{\text{3DGS}})$     \Comment{\textit{Main}.~Sec.~\ref{BDS}, Eq.~\eqref{eq:sdf_normal}}

\EndIf

\State Update $\mathcal{F} \leftarrow \arg\min_{\mathcal{F}} \mathcal{L}_{\text{SDF}}$ 
\Comment{\textit{Suppl}.~Sec.~\ref{Training_objective}, Eq.~\eqref{eq:supp_LSDF}}

\EndProcedure
\end{algorithmic}
\end{algorithm}

\begin{algorithm}[t]
\caption{\textbf{Step 2: Normal-guided Gaussian Alignment (NGA)}}
\label{alg:nga}
\small
\begin{algorithmic}[1]
\Procedure{NGA}{$\mathcal{G},\mathcal{F};\ S_{\text{NGA}}$}
\State \textbf{Freeze} $\mathcal{F}$

\State $\mathbf{d}_{\text{3DGS}} \leftarrow \textsc{RenderDepth}(\mathcal{G})$
\State $(\mathbf{d}_{\text{SDF}},\mathbf{n}_{\text{SDF}}) \leftarrow \textsc{RenderSDF}(\mathcal{F})$

\If{$\Ldep \in S_{\text{NGA}}$}
    \State Compute $\Ldep(\mathbf{d}_{\text{3DGS}}, \mathbf{d}_{\text{SDF}})$     \Comment{\textit{Suppl}.~Sec.~\ref{Training_objective}, Eq.~\eqref{eq:supp_Ldep}}
\EndIf
\If{$\LnorG \in S_{\text{NGA}}$}
    \State $\mathbf{n}^{\text{AB}}_{\text{3DGS}} \leftarrow \textsc{AlphaBlendedNormal}(\mathcal{G})$ 
    \Comment{\textit{Main.}~Fig.~\ref{fig:normal_comparison}~(A)}
    \State Compute $\LnorG(\mathbf{n}^{\text{AB}}_{\text{3DGS}},\mathbf{n}_{\text{SDF}})$     \Comment{\textit{Main}.~Sec.~\ref{BDS}, Eq.~\eqref{eq:3DGS_normal}}
\EndIf

\State Update $\mathcal{G} \leftarrow \arg\min_{\mathcal{G}} \mathcal{L}_{\text{3DGS}}$ 
\Comment{\textit{Suppl}.~Sec.~\ref{Training_objective}, Eq.~\eqref{eq:supp_L3DGS}}
\EndProcedure
\end{algorithmic}
\end{algorithm}

\subsection{Training Schedule}
Our training follows the two-stage pipeline summarized in Alg.~\ref{alg:corea_train}.
We first warm up 3DGS for 15k iterations, and then perform geometric coupling for 30k iterations, followed by inverse PBR for 10k iterations.

\noindent\textbf{Stage 1: Geometric coupling.}
Each iteration applies an alternating update scheme: we run DSA (Alg.~\ref{alg:dsa}) while freezing 3DGS and updating the SDF, and then run NGA (Alg.~\ref{alg:nga}) while freezing the SDF and updating 3DGS (Tab.~\ref{tab:ablation_alignment}).
For the first 10k iterations, we use depth consistency for coarse alignment.
We then enable pixel-wise normal consistency for the remaining 20k iterations to refine fine-scale geometry.
We activate DDC after 10k iterations and apply it every 100 iterations to regulate Gaussian splitting.

\noindent\textbf{Stage 2: Inverse PBR.}
We fix the learned geometry and normals of the Gaussians and optimize BRDF parameters and lighting for 10k iterations using $\mathcal{L}_{\text{PBR}}$ (Eq.~\eqref{PBRloss}), following inverse rendering setup of R3DG.

\subsection{Hyperparameter Settings}
In our framework, depth and normal consistency terms are shared by both the SDF and relightable 3DGS branches. 
\(\{\lambda_{\text{depth}}, \lambda_{\text{normal}}\}\) are set to \(\{0.01, 0.001\}\), where the normal weight is kept relatively small since normal gradients tend to dominate optimization, which often leads to unstable training.
For the SDF branch, we employ image reconstruction, Eikonal, and curvature regularization. 
\(\{\lambda_1^{\text{SDF}}, \lambda_{\text{eik}}, \lambda_{\text{curv}}\}\) are set to \(\{1.0, 0.1, 0.05\}\), following the configuration of GSDF~\cite{GSDF}.
For the relightable 3DGS branch, we adopt the weights from the first stage setup of R3DG~\cite{R3DG}. 
\(\{\lambda_1^{\text{3DGS}}, \lambda_{\text{ssim}}, \lambda_{O}, \lambda_{u}\}\) are set to \(\{0.8, 0.2, 0.01, 0.01\}\).
The DDC weight is set to \(\alpha = 0.2\), which balances photometric and geometric gradients during Gaussian densification, as described in Sec.~\ref{DDC_a_ablation}.

After geometry optimization, we fix the learned geometry and normals of the Gaussians and optimize BRDF parameters and lighting in the second stage. 
\(\{\lambda_1, \lambda_{\text{ssim}}, \lambda_{\text{light}}, \lambda_{\text{brdf}}\}\) are set to \(\{0.8, 0.2, 0.0001, 0.01\}\), following the second-stage configuration of R3DG.

\subsection{Complexity and Runtime}
All experiments were conducted on a single NVIDIA A6000 GPU.
On the DTU dataset, COREA requires approximately 2.4 hours of training per scene.
We report peak GPU memory usage, and COREA consumes roughly 24 GB on average during DTU training.
The implementation is built on PyTorch 2.3 and CUDA 12.1 within our unified pipeline for joint SDF and relightable 3DGS optimization.

\input{figs/A_ablation}

\section{Ablation Studies}

\subsection{Trade-off Analysis of $\alpha$ in DDC}
\label{DDC_a_ablation}
We analyze how the DDC weight $\alpha$ affects rendering performance and Gaussian splitting behavior.
To this end, we vary $\alpha \in \{0, 0.2, 0.5, 0.7, 1.0\}$ and measure three quantities: PSNR for SH-based NVS and inverse PBR, the number of Gaussians normalized by the count at $\alpha=0$, and the cosine similarity between the principal eigenvectors of the image-only splitting matrix $S_{\text{image}}$ and the combined matrix $S_{\text{total}}$ (Eq.~\eqref{eq:splitting_total}).
As shown in Fig.~\ref{fig:ablation_eigenvector}, the cosine similarity rapidly decreases once $\alpha \ge 0.5$, indicating that the splitting direction gradually shifts from an image-driven to a normal-driven direction.
This shift is accompanied by a slight increase in the relative number of Gaussians, suggesting that the normal loss induces additional splitting for fine geometric refinement.
While the SH-based NVS PSNR remains stable for all $\alpha$, the inverse PBR performance shows a mild improvement, peaking at $\alpha=0.2$. 
For larger $\alpha$, the rendering quality slightly decreases, implying that excessive normal-driven splitting destabilizes geometry learning.

These observations indicate that the DDC weight $\alpha$ controls the balance between photometric and geometric gradients, thereby regulating geometric fidelity and Gaussian growth.
Based on this empirical analysis, we adopt $\alpha = 0.2$ as the default setting, which provides the best balance between geometric fidelity, rendering quality, and controlled Gaussian growth across all experiments.

\subsection{Dual-Density Control for Efficient Gaussian Densification}
To further validate the effect of DDC on Gaussian growth, we conduct an ablation study on on the DTU dataset, as shown in Tab.~\ref{tab:DDC_ablation2}, which reports per-scene Gaussian counts.
With the Adaptive Density Control (ADC) used in vanilla 3DGS, the accumulated pixel-wise gradients from image and normal losses frequently trigger excessive Gaussian splitting, resulting in inefficient over-densification.
In contrast, DDC reduces the Gaussian count by 17\% on average while preserving rendering fidelity across all scenes.

By jointly considering photometric and geometric gradients, DDC triggers densification only when the combined gradients indicate meaningful refinement, effectively suppressing redundant growth.
This produces geometry-aware densification that remains compact and geometrically consistent, significantly reducing memory usage and improving the computational efficiency of inverse PBR, since fewer Gaussians participate in ray sampling and shading evaluations.
The resulting representation is lightweight yet geometrically accurate, supporting high-quality rendering and stable relighting.

\input{tab/ablation_DDC}

\input{figs/additional_results}

\section{More experiments}

\subsection{Per-scene Results}
We present additional qualitative results in Fig.~\ref{fig:other_results}, which demonstrate the visual performance of our method across diverse scenes. 
Per-scene quantitative results for SH-based NVS, surface reconstruction (SR), and inverse PBR are provided in Tabs.~\ref{tab:dtu_per_scenes_NVS}-\ref{tab:TnT_per_scenes_PBR}.

\begin{figure}[t]
    \centering    
    \includegraphics[width=\linewidth]{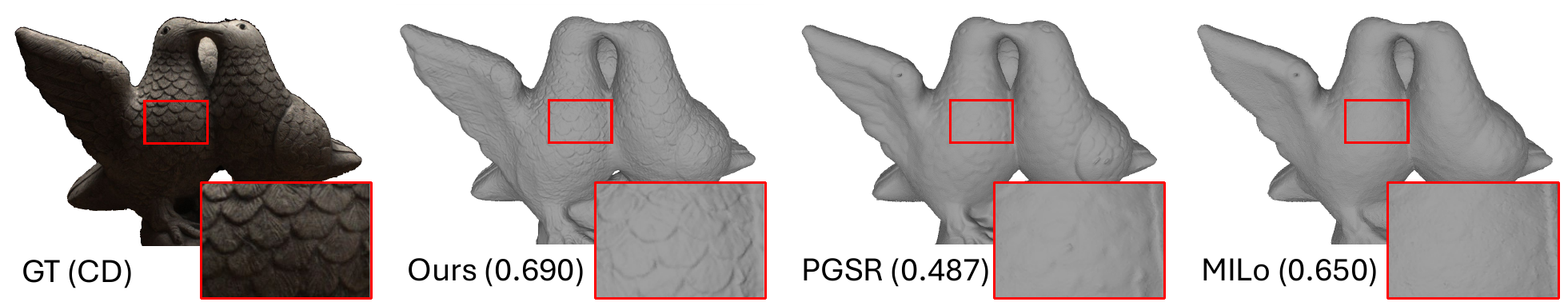}
    \caption{
    \textbf{Qualitative comparison of surface reconstruction under different CD.}
    We compare COREA with PGSR and MILo, which achieve the best and second-best CD in Tab.~\ref{tab:cat7_dtu_sota}, on an example DTU scene.
    Although COREA shows a higher CD (values in parentheses; lower is better), the zoomed-in region reveals that PGSR and MILo produce smoother surfaces with less distinct local patterns, whereas COREA preserves sharper surface structures.
    This suggests that CD does not fully reflect fine-scale surface details that affect local geometry and normal fidelity.
    }
    \label{fig:more_mesh_oriented}
\end{figure}

\begin{table}[t]
\centering
\caption{
\textbf{Additional comparison to mesh-oriented baselines on DTU.}
We report NVS image metrics (PSNR/SSIM/LPIPS) and surface reconstruction (SR) accuracy (CD) under a unified protocol for representative mesh-oriented 3DGS methods.
Most baselines focus on surface reconstruction and do not support inverse PBR, while GS-ROR$^2$ supports inverse PBR but does not provide SH-based NVS.
For a consistent image-quality reference, we report image metrics from each method's supported rendering mode (SH-based NVS when available, otherwise PBR rendering).
\textit{COREA is the only unified framework in this table that simultaneously supports SH-based NVS, surface reconstruction, and inverse PBR.}
}
\label{tab:cat7_dtu_sota}
\scalebox{0.8}{
\begin{tabular}{l|ccc|ccc|c}
\toprule
\textbf{Method} & \textbf{SH} & \textbf{SR} & \textbf{PBR} &
PSNR$\uparrow$ & SSIM$\uparrow$ & LPIPS$\downarrow$ &
CD$\downarrow$ \\
\midrule
2DGS (SIGGRAPH’24)       & \cmark & \cmark & \xmark & \cellcolor{best}34.65 & \cellcolor{second}0.939 & 0.165 & 0.801 \\
GOF (TOG’24)        & \cmark & \cmark & \xmark & 28.70 & 0.909 & \cellcolor{second}0.124 & 0.816 \\
RaDe-GS (Arxiv’24)    & \cmark & \cmark & \xmark & 28.99 & 0.917 & 0.158 & 0.677 \\
PGSR (TVCG’24)       & \cmark & \cmark & \xmark & 28.85 & 0.910 & 0.178 & \cellcolor{best}0.545 \\
MILo (TOG’24)       & \cmark & \cmark & \xmark & 29.45 & 0.919 & 0.159 & \cellcolor{second}0.673 \\
GS-ROR$^2$ (TOG’25) & \xmark & \cmark & \cmark & 19.76 & 0.812 & 0.219 & 1.562 \\
\midrule
\textbf{COREA (Ours)} & \cmark & \cmark & \cmark & \cellcolor{second}32.27 & \cellcolor{best}0.955 & \cellcolor{best}0.094 & 0.824 \\
\bottomrule
\end{tabular}
}
\end{table}

\subsection{Surface Reconstruction Comparison} 
\label{sec:more_mesh_oriented}
To provide a broader empirical context, we additionally compare against representative mesh-oriented 3DGS methods~\cite{2DGS,GaussianOpacityFields,Rade-gs,Pgsr,Milo,GSROR2} on DTU under a unified evaluation protocol.
These methods primarily target surface reconstruction quality.
Accordingly, we report image metrics and CD under each method's supported rendering mode to provide a consistent comparison across methods with different rendering capabilities.
Tab.~\ref{tab:cat7_dtu_sota} summarizes the results.
COREA is the only unified framework in this comparison that simultaneously supports SH-based NVS, surface reconstruction, and inverse PBR.
While COREA is not optimized solely for surface reconstruction, it maintains competitive CD on DTU, achieves strong NVS quality, and additionally supports inverse PBR.

Fig.~\ref{fig:more_mesh_oriented} highlights a qualitative discrepancy between CD and perceived surface detail.
Although COREA has a higher CD than PGSR and MILo, the meshes reconstructed by these methods appear overly smooth in the magnified region, with local surface patterns becoming less distinct.
\textit{In contrast, COREA better preserves fine local surface structures that influence normal fidelity, indicating that a lower CD does not necessarily imply higher fidelity in fine-scale geometry.~\cite{RNb-NeuS2}}

\section{Limitations}
\label{sec:limitations}
\begin{figure}[t]
    \centering
    \includegraphics[width=\linewidth]{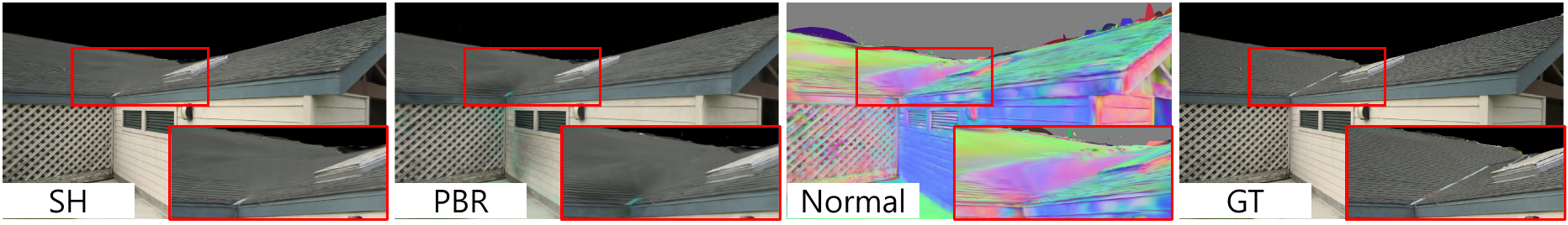}
    \caption{
    \textbf{Failure cases under low visibility and in thin-structure regions.}
    When observations are sparse or the geometry contains thin structures, both Gaussian coverage and SDF sampling can become unreliable.
    This may lead to over-smoothed local geometry and unstable relighting results.
    }
    \label{fig:limitations_sparse}
\end{figure}

Our framework relies on cross-representation geometric consistency between an SDF and relightable 3DGS, which can become fragile in geometrically ambiguous regions.
In sparse-view regions, depth and pixel-wise depth gradients can be weak or ambiguous due to limited visibility and depth discontinuities.
In such cases, early alignment errors may trigger a cascading feedback loop: Gaussians may collapse or become overly sparse, SDF ray sampling becomes biased around inaccurate depth anchors, and the local geometry progressively blurs.
This degradation is particularly visible under relighting, where small normal inaccuracies can amplify shading instability (Fig.~\ref{fig:limitations_sparse}).

A promising direction is to incorporate pixel-weighted Gaussian representations such as Pixel-GS~\cite{pixel_gs}, which may enable more reliable densification under sparse initial point clouds and improve Gaussian formation in low-visibility regions.
This may further stabilize relighting in challenging cases.

\newpage
\input{tab/ablation_DTU_NVS}
\input{tab/ablation_DTU_CD}
\input{tab/ablation_DTU_PBR}
\input{tab/ablation_TnT_NVS}
\input{tab/ablation_TnT_PBR}

%% file: figs/A_ablation.tex
\begin{figure}[!t]
  \centering
  \includegraphics[width=0.7\columnwidth]{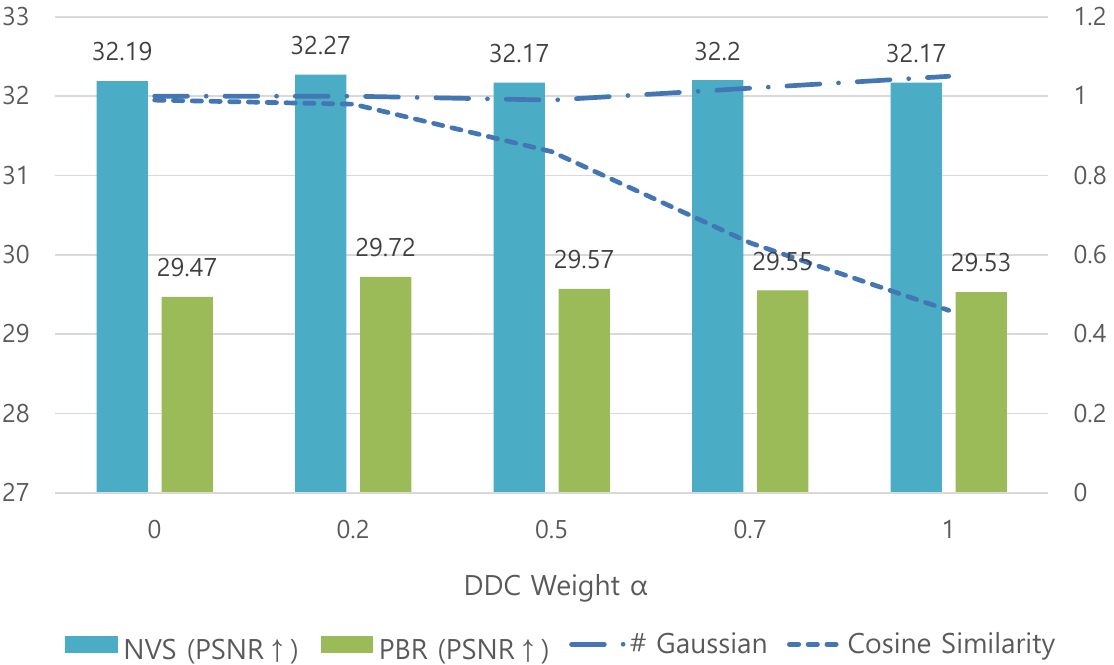}
  \caption{
    \textbf{Effect of the DDC weight $\alpha$ on rendering quality and splitting dynamics.}
    The bar plots show the PSNR for SH-based NVS and inverse PBR, and the line plot shows the cosine similarity between the principal eigenvectors of $S_{\text{image}}$ and $S_{\text{total}}$ (Eq.~\eqref{eq:splitting_total}).
    The relative number of Gaussians (normalized with respect to $\alpha{=}0$) shows how the Gaussian count changes with $\alpha$.   
    As $\alpha$ increases, the cosine similarity drops sharply for $\alpha \ge 0.5$, indicating a shift from image-driven to normal-driven splitting.
    The inverse PBR PSNR peaks at $\alpha{=}0.2$, showing the best balance between geometric fidelity and controlled Gaussian growth.
  }
\label{fig:ablation_eigenvector}
\end{figure}

%% file: tab/ablation_DDC.tex
\begin{table*}[!t]
\caption{
\textbf{Comparison between Adaptive and Dual-Density Control.}
All values indicate the number of Gaussians in millions (M).
The original Adaptive Density Control (ADC) in 3DGS 
tends to over-split Gaussians, whereas our Dual-Density Control (DDC) effectively suppresses redundant splitting through a dual-loss scheme that jointly considers image and normal gradients.
By regulating the generation of split Gaussians, DDC achieves geometry-aware densification and reduces the overall Gaussian count by an average of 17\%, significantly lowering memory usage and training overhead.
}
\centering
\scriptsize
\resizebox{\textwidth}{!}{%
\begin{tabular}{lcccccccccccccccc}
\toprule
\textbf{\#Gaussians$\downarrow$} &
\textbf{24} & \textbf{37} & \textbf{40} & \textbf{55} & \textbf{63} & \textbf{65} & \textbf{69} & \textbf{83} &
\textbf{97} & \textbf{105} & \textbf{106} & \textbf{110} & \textbf{114} & \textbf{118} & \textbf{122} & \textbf{Avg.} \\
\midrule
ADC
& 0.93 & 1.21 & 1.78 & 1.14 & 0.19 & 0.21 & 0.34 & 0.13 & 0.51 & 0.32 & 0.39 & 0.16 & 0.37 & 0.27 & 0.24 & \textbf{0.54} \\
DDC (Ours) 
& 0.89 & 1.18 & 1.65 & 1.08 & 0.19 & 0.21 & 0.36 & 0.13 & 0.50 & 0.32 & 0.37 & 0.16 & 0.36 & 0.26 & 0.24 & \textbf{0.45 (17\%↓)} \\
\bottomrule
\end{tabular}}
\label{tab:DDC_ablation2}
\end{table*}

%% file: figs/additional_results.tex
\begin{figure*}[t]
    \centering
    \includegraphics[width=0.98\textwidth]{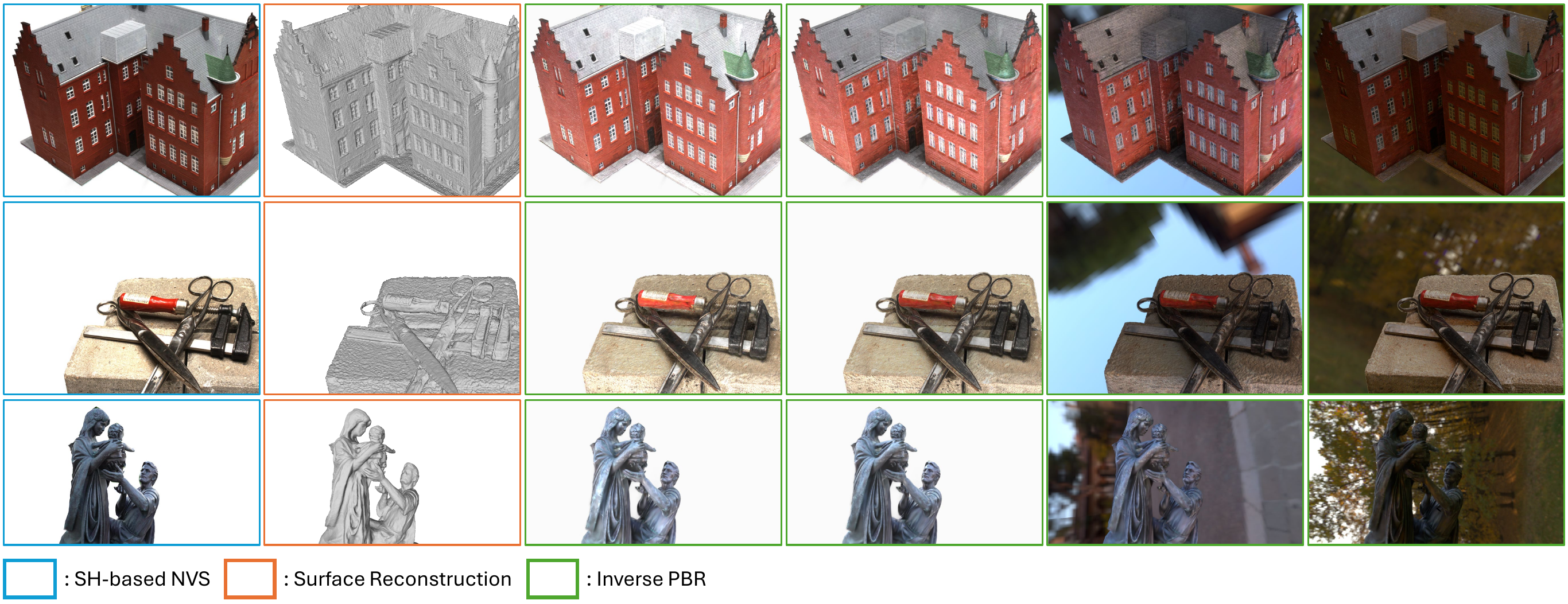}
    \caption{
    \textbf{Additional Qualitative Results.} 
    We show that \textbf{COREA} jointly supports SH-based NVS (blue), surface reconstruction (orange), and inverse PBR (green) within a unified three-tasks framework.
    Each row presents a different real-world scene, where COREA produces sharp novel views, accurate surface geometry, and stable relighting under white and HDR environment lighting, with clean object-background separation.
    These results highlight the effectiveness of the geometric coupling between an SDF and relightable 3DGS, which provides a shared geometric foundation across all three tasks.
    }
    \label{fig:other_results}
\end{figure*}

%% file: tab/ablation_DTU_NVS.tex
\begin{table*}[t]
  \centering
  \small
  \caption{Quantitative results for SH-based NVS on the DTU dataset.}
  \label{tab:dtu_per_scenes_NVS}

    \resizebox{\textwidth}{!}{%
    \begin{tabular}{lcc|ccccccccccccccc}
      \toprule
      \textbf{PSNR$\uparrow$} & SR & PBR &
      \textbf{24} & \textbf{37} & \textbf{40} & \textbf{55} & \textbf{63} &
      \textbf{65} & \textbf{69} & \textbf{83} & \textbf{97} & \textbf{105} &
      \textbf{106} & \textbf{110} & \textbf{114} & \textbf{118} & \textbf{122} \\
      \midrule
    3DGS & \xmark & \xmark &
      23.63 & 21.45 & 23.29 & 23.66 & 28.36 &
      28.35 & 27.64 & 30.30 & 24.43 & 27.55 &
      28.82 & 28.73 & 27.18 & 32.28 & 32.06 \\

    GaussianShader & \xmark & \cmark &
      22.28 & 9.61 & 20.57 & 23.09 & 16.54 &
      21.98 & 24.51 & 23.89 & 17.89 & 23.52 &
      19.86 & 25.86 & 22.09 & 29.24 & 29.72 \\

    GS-IR & \xmark & \cmark &
      23.86 & 21.55 & 23.18 & 24.66 & 28.74 &
      28.32 & 28.09 & \cellcolor{third}30.43 & 24.45 & 27.65 &
      29.45 & \cellcolor{third}29.65 & 27.33 & \cellcolor{third}32.86 & 33.07 \\

    R3DG & \xmark & \cmark &
      \cellcolor{best}28.54 & \cellcolor{best}25.41 & \cellcolor{best}27.62 & \cellcolor{best}32.98 & \cellcolor{second}31.39 &
      \cellcolor{second}31.82 & \cellcolor{third}29.60 & \cellcolor{second}32.43 &
      \cellcolor{second}27.04 & \cellcolor{third}30.01 &
      \cellcolor{best}35.51 & \cellcolor{best}33.61 & \cellcolor{second}31.16 & \cellcolor{second}37.98 & \cellcolor{second}37.79 \\

    SVG-IR & \xmark & \cmark &
      23.80 & 21.94 & 23.92 & 25.62 & 28.50 &
      27.14 & \cellcolor{best}29.78 & 30.94 & 25.65 & 27.74 &
      \cellcolor{third}30.21 & 30.98 & \cellcolor{third}28.32 & 32.44 & \cellcolor{third}33.33 \\

    \midrule

    GOF & \cmark & \xmark &
      24.18 & \cellcolor{third}22.50 & \cellcolor{third}24.25 & 24.12 & \cellcolor{third}30.53 &
      28.35 & 26.40 & \cellcolor{best}33.57 & \cellcolor{third}25.79 & \cellcolor{best}30.14 &
      28.84 & 28.67 & 25.52 & 31.14 & 30.98 \\

    GSDF & \cmark & \xmark &
      \cellcolor{third}24.87 & 22.09 & 23.84 & \cellcolor{third}25.74 & 28.48 &
      \cellcolor{third}29.18 & 26.81 & 25.45 & 24.92 & 27.74 &
      29.80 & 29.51 & 25.04 & 32.18 & 31.12 \\

    GS-pull & \cmark & \xmark &
      23.87 & 21.75 & 23.91 & 24.88 & 28.09 &
      27.19 & 27.70 & 29.69 & 24.73 & 26.83 &
      29.59 & 28.41 & 27.57 & 32.33 & 32.24 \\


    \textbf{COREA (Ours)} & \cmark & \cmark &
      \cellcolor{second}28.47 & \cellcolor{second}25.35 & \cellcolor{second}26.95 & \cellcolor{second}32.89 & \cellcolor{best}31.49 &
      \cellcolor{best}31.87 & \cellcolor{second}29.68 & \cellcolor{second}32.43 &
      \cellcolor{best}27.38 & \cellcolor{second}30.11 &
      \cellcolor{second}35.46 & \cellcolor{second}33.48 & \cellcolor{best}31.34 & \cellcolor{best}38.17 & \cellcolor{best}38.13 \\
      
      \bottomrule
    \end{tabular}
    }
    
    \vspace{2mm}

    \resizebox{\textwidth}{!}{%
    \begin{tabular}{lcc|ccccccccccccccc}
      \toprule
      \textbf{SSIM$\uparrow$} & SR & PBR &
      \textbf{24} & \textbf{37} & \textbf{40} & \textbf{55} & \textbf{63} &
      \textbf{65} & \textbf{69} & \textbf{83} & \textbf{97} & \textbf{105} &
      \textbf{106} & \textbf{110} & \textbf{114} & \textbf{118} & \textbf{122} \\
      \midrule    
    3DGS         & \xmark & \xmark &
          0.728 & 0.720 & 0.674 & 0.802 & 0.910 &
          0.927 & 0.883 & 0.946 & 0.865 & 0.877 &
          0.870 & 0.906 & 0.847 & 0.911 & 0.925 \\
    
    GaussianShader & \xmark & \cmark &
          0.711 & 0.592 & 0.642 & 0.814 & 0.797 &
          0.915 & 0.801 & 0.869 & 0.760 & 0.816 &
          0.789 & 0.881 & 0.800 & 0.905 & 0.892 \\

    GS-IR        & \xmark & \cmark &
          0.730 & 0.725 & 0.665 & 0.809 & 0.929 &
          0.929 & 0.886 & 0.947 & 0.865 & 0.874 &
          0.866 & 0.911 & 0.847 & 0.912 & 0.928 \\
    
    R3DG         & \xmark & \cmark &
          \cellcolor{second}0.921 & \cellcolor{best}0.894 & \cellcolor{second}0.891 & \cellcolor{second}0.966 & \cellcolor{second}0.961 &
          \cellcolor{second}0.961 & \cellcolor{second}0.935 & \cellcolor{second}0.969 & \cellcolor{second}0.932 & \cellcolor{second}0.944 &
          \cellcolor{second}0.957 & \cellcolor{third}0.954 & \cellcolor{best}0.937 & \cellcolor{second}0.966 & \cellcolor{second}0.971 \\
    
    SVG-IR & \xmark & \cmark &
      \cellcolor{third}0.869 & \cellcolor{second}0.859 & \cellcolor{third}0.836 & 0.901 & 0.951 &
      \cellcolor{third}0.954 & \cellcolor{third}0.926 & 0.960 & \cellcolor{third}0.918 & \cellcolor{third}0.926 &
      \cellcolor{third}0.942 & \cellcolor{second}0.959 & \cellcolor{second}0.928 & \cellcolor{third}0.957 & \cellcolor{third}0.963 \\
    
    \midrule

    GOF & \cmark & \xmark &
      0.749 & 0.741 & 0.700 & 0.825 & \cellcolor{third}0.941 &
      0.926 & 0.893 & \cellcolor{third}0.965 & 0.878 & 0.900 &
      0.880 & 0.925 & 0.846 & 0.913 & 0.931 \\
    
    GSDF         & \cmark & \xmark &
          0.774 & 0.741 & 0.697 & \cellcolor{third}0.847 & 0.932 &
          0.939 & 0.844 & 0.889 & 0.881 & 0.884 &
          0.886 & 0.913 & 0.820 & 0.908 & 0.910 \\
    
    GS-pull      & \cmark & \xmark &
          0.750 & \cellcolor{third}0.745 & 0.691 & 0.818 & 0.926 &
          0.927 & 0.888 & 0.941 & 0.872 & 0.870 &
          0.882 & 0.907 & \cellcolor{third}0.854 & 0.915 & 0.928 \\
    

    \textbf{COREA(Ours)} & \cmark & \cmark &
          \cellcolor{best}0.922 & \cellcolor{best}0.894 & \cellcolor{best}0.930 & \cellcolor{best}0.976 & \cellcolor{best}0.967 &
          \cellcolor{best}0.968 & \cellcolor{best}0.951 & \cellcolor{best}0.970 & \cellcolor{best}0.957 & \cellcolor{best}0.955 &
          \cellcolor{best}0.964 & \cellcolor{best}0.961 & \cellcolor{best}0.937 & \cellcolor{best}0.967 & \cellcolor{best}0.973 \\

      \bottomrule
      
    \end{tabular}
    }
    
    \vspace{2mm}
    
    \resizebox{\textwidth}{!}{%
    \begin{tabular}{lcc|ccccccccccccccc}
      \toprule
      \textbf{LPIPS$\downarrow$} & SR & PBR &
      \textbf{24} & \textbf{37} & \textbf{40} & \textbf{55} & \textbf{63} &
      \textbf{65} & \textbf{69} & \textbf{83} & \textbf{97} & \textbf{105} &
      \textbf{106} & \textbf{110} & \textbf{114} & \textbf{118} & \textbf{122} \\
      \midrule

    3DGS         & \xmark & \xmark &
          0.289 & 0.224 & 0.325 & 0.174 & 0.129 &
          0.122 & 0.222 & 0.113 & 0.183 & 0.201 &
          0.195 & 0.177 & 0.219 & 0.167 & 0.125 \\
    
    GaussianShader & \xmark & \cmark &
          0.264 & 0.318 & 0.304 & 0.180 & 0.206 &
          0.144 & 0.244 & 0.143 & 0.229 & 0.230 &
          0.216 & 0.193 & 0.246 & 0.172 & 0.137 \\

    GS-IR        & \xmark & \cmark &
          0.246 & 0.203 & 0.282 & 0.152 & 0.120 &
          0.116 & 0.207 & 0.102 & 0.173 & 0.179 &
          0.179 & 0.164 & 0.202 & 0.148 & 0.111 \\
    
    R3DG         & \xmark & \cmark &
          \cellcolor{third}0.095 & \cellcolor{third}0.105 & \cellcolor{third}0.157 & \cellcolor{third}0.056 & \cellcolor{third}0.073 &
          0.080 & 0.146 & 0.075 & 0.113 & 0.109 &
          0.100 & 0.121 & \cellcolor{third}0.105 & 0.091 & 0.063 \\
    
    SVG-IR & \xmark & \cmark &
      \cellcolor{second}0.093 & \cellcolor{best}0.079 & \cellcolor{best}0.127 & \cellcolor{second}0.055 & \cellcolor{best}0.035 &
      \cellcolor{best}0.037 & \cellcolor{best}0.057 & \cellcolor{second}0.027 & \cellcolor{best}0.053 & \cellcolor{second}0.050 &
      \cellcolor{best}0.046 & \cellcolor{second}0.050 & \cellcolor{best}0.065 & \cellcolor{best}0.035 & \cellcolor{best}0.023 \\
    
    \midrule
    
    GOF & \cmark & \xmark &
      0.128 & 0.111 & 0.160 & 0.079 & \cellcolor{best}0.035 &
      \cellcolor{second}0.069 & \cellcolor{second}0.079 & \cellcolor{best}0.021 & \cellcolor{second}0.065 & \cellcolor{best}0.048 &
      \cellcolor{second}0.063 & \cellcolor{best}0.048 & \cellcolor{second}0.088 & \cellcolor{second}0.053 & \cellcolor{second}0.041 \\
    
    GSDF         & \cmark & \xmark &
          0.243 & 0.206 & 0.294 & 0.152 & 0.124 &
          0.115 & 0.225 & 0.136 & 0.170 & 0.186 &
          0.179 & 0.179 & 0.218 & 0.152 & 0.121 \\
    
    GS-pull      & \cmark & \xmark &
          0.321 & 0.245 & 0.351 & 0.187 & 0.144 &
          0.136 & 0.250 & 0.126 & 0.205 & 0.226 &
          0.212 & 0.194 & 0.248 & 0.186 & 0.141 \\
        

    \textbf{COREA(Ours)} & \cmark & \cmark &
          \cellcolor{best}0.088 & \cellcolor{second}0.099 & \cellcolor{second}0.134 & \cellcolor{best}0.050 & \cellcolor{second}0.072 &
          \cellcolor{third}0.077 & \cellcolor{third}0.142 & \cellcolor{third}0.065 & \cellcolor{third}0.100 & \cellcolor{third}0.099 &
          \cellcolor{third}0.097 & \cellcolor{third}0.113 & \cellcolor{third}0.105 & \cellcolor{third}0.085 & \cellcolor{third}0.056 \\
      \bottomrule
    \end{tabular}
    }
\end{table*}

%% file: tab/ablation_DTU_CD.tex
\begin{table*}[t]
  \centering
  \small
  \caption{Quantitative results for surface reconstruction (CD) on the DTU dataset.}
  \resizebox{\textwidth}{!}{%
    \begin{tabular}{lcc|ccccccccccccccc}
      \toprule
      \textbf{CD$\downarrow$} & SR & PBR &
      \textbf{24} & \textbf{37} & \textbf{40} & \textbf{55} & \textbf{63} &
      \textbf{65} & \textbf{69} & \textbf{83} & \textbf{97} & \textbf{105} &
      \textbf{106} & \textbf{110} & \textbf{114} & \textbf{118} & \textbf{122} \\
      \midrule
      
    GOF & \cmark & \xmark &
      \cellcolor{best}0.45 & \cellcolor{best}0.80 & \cellcolor{second}0.56 & \cellcolor{second}0.44 & 1.50 &
      \cellcolor{third}1.18 & \cellcolor{third}0.77 & \cellcolor{best}1.16 & \cellcolor{third}1.42 & \cellcolor{best}0.57 &
      \cellcolor{best}0.58 & \cellcolor{third}1.08 & \cellcolor{third}0.54 & \cellcolor{best}0.49 & \cellcolor{best}0.42 \\
      
      GSDF      & \cmark & \xmark &
      1.03 & 0.90 & \cellcolor{best}0.47 & \cellcolor{second}0.44 & \cellcolor{second}1.32 &
      \cellcolor{best}0.84 & \cellcolor{second}0.76 & \cellcolor{third}1.59 & \cellcolor{best}1.33 & \cellcolor{third}0.78 &
      \cellcolor{second}0.60 & 1.30 & \cellcolor{second}0.39 & \cellcolor{second}0.52 & \cellcolor{third}0.56 \\
      
      GS-pull   & \cmark & \xmark &
      \cellcolor{second}0.62 & \cellcolor{second}0.71 & 1.45 & \cellcolor{third}0.63 & \cellcolor{best}0.90 &
      1.30 & 0.88 & 6.26 & 1.45 & 2.11 &
      0.80 & \cellcolor{best}0.99 & 0.55 & 0.66 & 12.83 \\
    
    GS-ROR & \cmark & \cmark &
      3.16 & 4.60 & 3.51 & 0.26 & 2.06 &
      1.95 & 2.76 & 2.24 & 3.10 & 1.90 &
      1.87 & 3.47 & 3.24 & 2.11 & 2.06 \\
      
      \textbf{COREA(Ours)} & \cmark & \cmark &
      \cellcolor{third}0.77 & \cellcolor{third}0.87 & \cellcolor{third}0.58 & \cellcolor{best}0.38 & \cellcolor{third}1.41 &
      \cellcolor{second}0.94 & \cellcolor{best}0.70 & \cellcolor{second}1.49 & \cellcolor{second}1.36 & \cellcolor{second}0.77 &
      \cellcolor{third}0.69 & \cellcolor{second}1.02 & \cellcolor{best}0.36 & \cellcolor{third}0.53 & \cellcolor{second}0.49 \\
      \bottomrule
    \end{tabular}
  }
  \label{tab:dtu_cd}
\end{table*}

%% file: tab/ablation_DTU_PBR.tex
\begin{table*}[!h]
  \centering
  \small
  \caption{Quantitative results for Inverse PBR on the DTU dataset.}
  \label{tab:dtu_per_scenes_PBR}

    \resizebox{\textwidth}{!}{%
    \begin{tabular}{lcc|ccccccccccccccc}
      \toprule
      \textbf{PSNR$\uparrow$} & SR & PBR &
      \textbf{24} & \textbf{37} & \textbf{40} &
      \textbf{55} & \textbf{63} & \textbf{65} & \textbf{69} & \textbf{83} &
      \textbf{97} & \textbf{105} & \textbf{106} & \textbf{110} & \textbf{114} & 
      \textbf{118} & \textbf{122} \\
      \midrule
      
    GaussianShader & \xmark & \cmark &
          19.56 & \cellcolor{second}23.21 & 17.83 & 15.76 & 23.18 &
          16.75 & 16.04 & 24.00 & 20.43 & 23.73 &
          17.35 & 17.59 & 16.76 & 18.36 & 18.05 \\

    GS-IR     & \xmark & \cmark &
          \cellcolor{third}21.06 & 20.46 & \cellcolor{third}21.73 & \cellcolor{third}22.77 & \cellcolor{third}25.40 &
          \cellcolor{third}23.53 & \cellcolor{second}26.07 & \cellcolor{second}29.24 & \cellcolor{third}23.24 & \cellcolor{third}24.61 &
          \cellcolor{third}27.16 & \cellcolor{second}27.77 & \cellcolor{third}25.04 & \cellcolor{third}27.58 & \cellcolor{third}27.48 \\
          
    R3DG      & \xmark & \cmark &
          \cellcolor{second}24.32 & \cellcolor{third}22.36 & \cellcolor{second}24.52 & \cellcolor{second}27.14 & \cellcolor{second}27.26 &
          \cellcolor{second}27.38 & \cellcolor{third}26.03 & \cellcolor{third}27.13 & \cellcolor{second}23.77 & \cellcolor{second}25.93 &
          \cellcolor{second}29.45 & \cellcolor{third}25.61 & \cellcolor{second}25.26 & \cellcolor{second}30.56 & \cellcolor{second}31.53 \\
          
    SVG-IR & \xmark & \cmark &
      \textit{OOM} & \textit{OOM} & \textit{OOM} & \textit{OOM} & \textit{OOM} &
      \textit{OOM} & \textit{OOM} & \textit{OOM} & \textit{OOM} & \textit{OOM} &
      \textit{OOM} & \textit{OOM} & \textit{OOM} & \textit{OOM} & \textit{OOM} \\
    
    GS-ROR & \cmark & \cmark &
      15.74 & 15.68 & 16.34 & 18.75 & 20.99 &
      15.84 & 22.01 & 26.47 & 20.54 & 22.47 &
      22.26 & 20.71 & 16.69 & 25.07 & 26.25 \\
      
    \textbf{COREA (Ours)}  & \cmark & \cmark &
          \cellcolor{best}26.98 & \cellcolor{best}25.48 & \cellcolor{best}27.64 & \cellcolor{best}30.25 & \cellcolor{best}30.28 &
          \cellcolor{best}30.50 & \cellcolor{best}28.31 & \cellcolor{best}31.16 & \cellcolor{best}25.82 & \cellcolor{best}29.40 &
          \cellcolor{best}32.20 & \cellcolor{best}29.89 & \cellcolor{best}27.70 & \cellcolor{best}33.38 & \cellcolor{best}34.58 \\
      \bottomrule
    \end{tabular}
    }

    \vspace{2mm}
    
    \resizebox{\textwidth}{!}{%
    \begin{tabular}{lcc|ccccccccccccccc}
      \toprule
      \textbf{SSIM$\uparrow$} & SR & PBR &
      \textbf{24} & \textbf{37} & \textbf{40} &
      \textbf{55} & \textbf{63} & \textbf{65} & \textbf{69} & \textbf{83} &
      \textbf{97} & \textbf{105} & \textbf{106} & \textbf{110} & \textbf{114} & 
      \textbf{118} & \textbf{122} \\
      \midrule

    GaussianShader & \xmark & \cmark &
          \cellcolor{third}0.827 & \cellcolor{third}0.843 & 0.733 & 0.563 & 0.897 &
          0.729 & 0.574 & 0.823 & 0.719 & 0.831 &
          0.614 & 0.527 & 0.652 & 0.598 & 0.599 \\

    GS-IR     & \xmark & \cmark &
          0.747 & 0.781 & \cellcolor{third}0.734 & 0.662 & 0.830 &
          0.715 & 0.762 & \cellcolor{second}0.952 & \cellcolor{second}0.892 & 0.602 &
          0.799 & 0.826 & 0.621 & 0.581 & 0.539 \\

    R3DG      & \xmark & \cmark &
          \cellcolor{second}0.869 & \cellcolor{second}0.850 & \cellcolor{second}0.850 &
          \cellcolor{second}0.932 & \cellcolor{third}0.917 & \cellcolor{second}0.946 &
          \cellcolor{second}0.896 & \cellcolor{third}0.948 & \cellcolor{third}0.890 &
          \cellcolor{second}0.908 & \cellcolor{second}0.932 & \cellcolor{second}0.913 &
          \cellcolor{second}0.911 & \cellcolor{second}0.944 & \cellcolor{second}0.953 \\
          
    SVG-IR & \xmark & \cmark &
      \textit{OOM} & \textit{OOM} & \textit{OOM} & \textit{OOM} & \textit{OOM} &
      \textit{OOM} & \textit{OOM} & \textit{OOM} & \textit{OOM} & \textit{OOM} &
      \textit{OOM} & \textit{OOM} & \textit{OOM} & \textit{OOM} & \textit{OOM} \\
    
    GS-ROR & \cmark & \cmark &
      0.724 & 0.751 & 0.712 & \cellcolor{third}0.822 & \cellcolor{second}0.918 &
      \cellcolor{third}0.894 & \cellcolor{third}0.869 & 0.940 & 0.859 & \cellcolor{third}0.880 &
      \cellcolor{third}0.877 & \cellcolor{third}0.867 & \cellcolor{third}0.817 & \cellcolor{third}0.908 & \cellcolor{third}0.926 \\
      
    \textbf{COREA (Ours)}  & \cmark & \cmark &
          \cellcolor{best}0.917 & \cellcolor{best}0.909 & \cellcolor{best}0.902 &
          \cellcolor{best}0.960 & \cellcolor{best}0.955 & \cellcolor{best}0.959 &
          \cellcolor{best}0.928 & \cellcolor{best}0.963 & \cellcolor{best}0.934 &
          \cellcolor{best}0.938 & \cellcolor{best}0.947 & \cellcolor{best}0.941 &
          \cellcolor{best}0.934 & \cellcolor{best}0.959 & \cellcolor{best}0.969 \\
      \bottomrule
    \end{tabular}
    }
    
    \vspace{2mm}
    
    \resizebox{\textwidth}{!}{%
    \begin{tabular}{lcc|ccccccccccccccc}
      \toprule
      \textbf{LPIPS$\downarrow$} & SR & PBR &
      \textbf{24} & \textbf{37} & \textbf{40} &
      \textbf{55} & \textbf{63} & \textbf{65} & \textbf{69} & \textbf{83} &
      \textbf{97} & \textbf{105} & \textbf{106} & \textbf{110} & \textbf{114} & 
      \textbf{118} & \textbf{122} \\
      \midrule
      
      GaussianShader & \xmark & \cmark &
      0.213 & 0.225 & 0.313 &
      0.519 & 0.208 & 0.458 &
      0.519 & 0.396 & 0.399 &
      0.345 & 0.455 & 0.497 &
      0.436 & 0.493 & 0.495 \\

      GS-IR     & \xmark & \cmark &
      \cellcolor{third}0.198 & \cellcolor{third}0.176 & \cellcolor{third}0.265 &
      0.217 & \cellcolor{third}0.154 & 0.205 &
      0.258 & \cellcolor{second}0.086 & \cellcolor{second}0.145 &
      0.231 & 0.208 & 0.200 &
      0.270 & 0.204 & 0.240 \\
      
      R3DG      & \xmark & \cmark &
      \cellcolor{second}0.154 & \cellcolor{second}0.142 & \cellcolor{second}0.200 &
      \cellcolor{second}0.089 & \cellcolor{second}0.105 & \cellcolor{second}0.097 &
      \cellcolor{second}0.196 & \cellcolor{third}0.104 & \cellcolor{third}0.154 &
      \cellcolor{second}0.149 & \cellcolor{second}0.133 & \cellcolor{second}0.166 &
      \cellcolor{second}0.136 & \cellcolor{second}0.118 & \cellcolor{second}0.082 \\
      
    SVG-IR & \xmark & \cmark &
      \textit{OOM} & \textit{OOM} & \textit{OOM} & \textit{OOM} & \textit{OOM} &
      \textit{OOM} & \textit{OOM} & \textit{OOM} & \textit{OOM} & \textit{OOM} &
      \textit{OOM} & \textit{OOM} & \textit{OOM} & \textit{OOM} & \textit{OOM} \\
    
    GS-ROR & \cmark & \cmark &
      0.314 & 0.241 & 0.329 & \cellcolor{third}0.175 & 0.149 &
      \cellcolor{third}0.180 & \cellcolor{third}0.238 & 0.113 & 0.204 & \cellcolor{third}0.211 &
      \cellcolor{third}0.193 & \cellcolor{third}0.199 & \cellcolor{third}0.249 & \cellcolor{third}0.175 & \cellcolor{third}0.125 \\
      
      \textbf{COREA (Ours)}  & \cmark & \cmark &
      \cellcolor{best}0.105 & \cellcolor{best}0.106 & \cellcolor{best}0.160 &
      \cellcolor{best}0.064 & \cellcolor{best}0.086 & \cellcolor{best}0.085 &
      \cellcolor{best}0.150 & \cellcolor{best}0.076 & \cellcolor{best}0.122 &
      \cellcolor{best}0.115 & \cellcolor{best}0.107 & \cellcolor{best}0.132 &
      \cellcolor{best}0.131 & \cellcolor{best}0.094 & \cellcolor{best}0.062 \\
      \bottomrule
    \end{tabular}
    }

\end{table*}

%% file: tab/ablation_TnT_NVS.tex
\begin{table*}[!t]
    \centering
    \caption{Quantitative results for SH-based NVS on the Tanks\&Temples dataset.}
    \label{tab:TnT_per_scenes_NVS}
    \renewcommand{\arraystretch}{1.1}
    \resizebox{\textwidth}{!}{%
        \begin{tabular}{lcc|ccc|ccc|ccc|ccc}
          \toprule
          & & & 
          \multicolumn{3}{c|}{\textbf{Barn}} &
          \multicolumn{3}{c|}{\textbf{Caterpillar}} & 
          \multicolumn{3}{c|}{\textbf{Family}} & 
          \multicolumn{3}{c}{\textbf{Truck}} \\ 
          \cmidrule(lr){4-6}\cmidrule(lr){7-9}\cmidrule(lr){10-12}\cmidrule(lr){13-15}
          \textbf{Method} & SR & PBR &
          PSNR$\uparrow$ & SSIM$\uparrow$ & LPIPS$\downarrow$ &
          PSNR$\uparrow$ & SSIM$\uparrow$ & LPIPS$\downarrow$ & 
          PSNR$\uparrow$ & SSIM$\uparrow$ & LPIPS$\downarrow$ & 
          PSNR$\uparrow$ & SSIM$\uparrow$ & LPIPS$\downarrow$ \\
          \midrule
          
          3DGS          & \xmark & \xmark & 
          27.53 & 0.879 & 0.158 & 
          24.45 & 0.891 & 0.133 & 
          32.34 & 0.963 & 0.050 & 
          25.51 & 0.908 & 0.119 \\
          
          GaussianShader     & \xmark & \cmark &
          21.97 & 0.878 & \cellcolor{third}0.137 &
          21.50 & 0.867 & 0.119 &
          20.47 & 0.904 & 0.088 &
          15.89 & 0.754 & 0.213 \\

          GS-IR         & \xmark & \cmark & 
          \cellcolor{second}28.28 & \cellcolor{second}0.900 & 0.141 & 
          \cellcolor{third}26.75 & 0.933 & \cellcolor{third}0.077 & 
          \cellcolor{third}32.65 & 0.965 & 0.041 & 
          \cellcolor{third}26.02 & \cellcolor{third}0.918 & \cellcolor{third}0.096 \\

          R3DG          & \xmark & \cmark & 
          \cellcolor{best}28.82 & \cellcolor{best}0.906 & \cellcolor{second}0.138 & 
          \cellcolor{best}27.92 & \cellcolor{best}0.949 & \cellcolor{second}0.067 & 
          \cellcolor{second}34.60 & \cellcolor{second}0.977 & \cellcolor{second}0.031 & 
          \cellcolor{second}26.34 & \cellcolor{second}0.933 & \cellcolor{second}0.081 \\
          
          SVG-IR      & \xmark & \cmark & 
          27.49 & \cellcolor{best}0.906 & \cellcolor{best}0.074 & 
          26.73 & \cellcolor{second}0.944 & \cellcolor{best}0.037 & 
          31.04 & \cellcolor{third}0.971 & \cellcolor{best}0.021 & 
          25.49 & \cellcolor{third}0.918 & \cellcolor{best}0.050 \\
          
          \midrule

          GOF      & \cmark & \xmark & 
          \textit{OOM} & \textit{OOM} & \textit{OOM} & 
          \textit{OOM} & \textit{OOM} & \textit{OOM} & 
          \textit{OOM} & \textit{OOM} & \textit{OOM} & 
          \textit{OOM} & \textit{OOM} & \textit{OOM} \\
          
          GSDF          & \cmark & \xmark & 
          \textit{OOM} & \textit{OOM} & \textit{OOM} & 
          \textit{OOM} & \textit{OOM} & \textit{OOM} & 
          \textit{OOM} & \textit{OOM} & \textit{OOM} & 
          \textit{OOM} & \textit{OOM} & \textit{OOM} \\
          
          GS-pull       & \cmark & \xmark & 
          26.25 & 0.847 & 0.206 & 
          22.80 & 0.870 & 0.158 & 
          31.38 & 0.957 & 0.060 & 
          24.71 & 0.897 & 0.143 \\

          
          \textbf{COREA(Ours)} & \cmark & \cmark & 
          \cellcolor{third}27.71 & \cellcolor{third}0.894 & 0.145 & 
          \cellcolor{second}27.75 & \cellcolor{third}0.943 & 0.083 & 
          \cellcolor{best}34.78 & \cellcolor{best}0.978 & \cellcolor{third}0.035 & 
          \cellcolor{best}26.76 & \cellcolor{best}0.934 & \cellcolor{third}0.096 \\
          \bottomrule
        \end{tabular}
    }
    \vspace{1pt}
\end{table*}

%% file: tab/ablation_TnT_PBR.tex
\begin{table*}[!t]
    \centering
    \caption{Quantitative results for Inverse PBR on the Tanks\&Temples dataset.}
    \label{tab:TnT_per_scenes_PBR}
    \renewcommand{\arraystretch}{1.1}
    \resizebox{\textwidth}{!}{%
        \begin{tabular}{lcc|ccc|ccc|ccc|ccc}
          \toprule
          & & & 
          \multicolumn{3}{c|}{\textbf{Barn}} &
          \multicolumn{3}{c|}{\textbf{Caterpillar}} & 
          \multicolumn{3}{c|}{\textbf{Family}} & 
          \multicolumn{3}{c}{\textbf{Truck}} \\ 
          \cmidrule(lr){4-6}\cmidrule(lr){7-9}\cmidrule(lr){10-12}\cmidrule(lr){13-15}
          \textbf{Method} & SR & PBR &
          PSNR$\uparrow$ & SSIM$\uparrow$ & LPIPS$\downarrow$ &
          PSNR$\uparrow$ & SSIM$\uparrow$ & LPIPS$\downarrow$ & 
          PSNR$\uparrow$ & SSIM$\uparrow$ & LPIPS$\downarrow$ & 
          PSNR$\uparrow$ & SSIM$\uparrow$ & LPIPS$\downarrow$ \\
          \midrule
          
          GaussianShader & \xmark & \cmark &
          18.72 & 0.816 & \cellcolor{third}0.202 &
          17.54 & 0.849 & 0.156 &
          20.61 & 0.888 & 0.103 &
          14.40 & 0.762 & 0.215 \\

          GS-IR     & \xmark & \cmark &
          \cellcolor{third}22.27 & 0.611 & 0.249 &
          21.45 & 0.390 & 0.191 &
          \cellcolor{third}28.05 & 0.924 & 0.079 &
          \cellcolor{third}20.08 & 0.682 & 0.225 \\
          
          R3DG      & \xmark & \cmark &
          \cellcolor{second}26.11 & \cellcolor{second}0.881 & \cellcolor{second}0.159 &
          \cellcolor{second}24.27 & \cellcolor{third}0.900 & \cellcolor{best}0.098 &
          \cellcolor{second}30.52 & \cellcolor{second}0.960 & \cellcolor{second}0.046 & 
          \cellcolor{second}24.35 & \cellcolor{second}0.906 & \cellcolor{second}0.108 \\

          SVG-IR      & \xmark & \cmark & 
          \textit{OOM} & \textit{OOM} & \textit{OOM} & 
          \textit{OOM} & \textit{OOM} & \textit{OOM} & 
          \textit{OOM} & \textit{OOM} & \textit{OOM} & 
          \textit{OOM} & \textit{OOM} & \textit{OOM} \\

          GS-ROR      & \cmark & \cmark & 
          22.20 & \cellcolor{third}0.825 & 0.252 & 
          \cellcolor{third}21.54 & \cellcolor{second}0.902 & \cellcolor{third}0.109 & 
          26.27 & \cellcolor{third}0.943 & \cellcolor{third}0.075 & 
          19.75 & \cellcolor{third}0.879 & \cellcolor{third}0.145 \\
          
          \textbf{COREA(Ours)} & \cmark & \cmark &
          \cellcolor{best}26.98 & \cellcolor{best}0.895 & \cellcolor{best}0.138 &
          \cellcolor{best}25.14 & \cellcolor{best}0.908 & \cellcolor{second}0.103 &
          \cellcolor{best}31.99 & \cellcolor{best}0.967 & \cellcolor{best}0.045 &
          \cellcolor{best}25.42 & \cellcolor{best}0.920 & \cellcolor{best}0.107 \\
          \bottomrule
        \end{tabular}
    }
    \vspace{1pt}
\end{table*}